\documentclass[journal]{IEEEtran}
\usepackage[mathscr]{eucal}
\usepackage[cmex10]{amsmath}
\usepackage{epsfig,epsf,psfrag}
\usepackage{amssymb,amsmath,amsthm,amsfonts,latexsym}
\usepackage{xcolor,url,overpic}
\usepackage{array}
\usepackage{verbatim}
\usepackage{bm}
\usepackage{bbm}
\usepackage{dsfont}
\usepackage{algorithmic}
\usepackage{algorithm}
\usepackage{verbatim}
\usepackage{textcomp}
\usepackage{mathrsfs}
\usepackage{epstopdf}
\usepackage{booktabs} 

\newcommand{\openone}{\leavevmode\hbox{\small1\normalsize\kern-.33em1}}

\catcode`~=11 \def\UrlSpecials{\do\~{\kern -.15em\lower .7ex\hbox{~}\kern .04em}} \catcode`~=13 

\allowdisplaybreaks[1]

\newcommand{\nn}{\nonumber}

\newcommand{\calA}{\mathcal{A}}

\newcommand{\calJ}{\mathcal{J}}

\newcommand{\calL}{\mathcal{L}}
\newcommand{\calM}{\mathcal{M}}
\newcommand{\calN}{\mathcal{N}}

\newcommand{\calP}{\mathcal{P}}

\newcommand{\calT}{\mathcal{T}}

\newcommand{\calW}{\mathcal{W}}
\newcommand{\calX}{\mathcal{X}}
\newcommand{\calY}{\mathcal{Y}}

\newcommand{\bA}{\mathbf{A}}

\newcommand{\bB}{\mathbf{B}}

\newcommand{\bE}{\mathbf{E}}

\newcommand{\bI}{\mathbf{I}}

\newcommand{\bM}{\mathbf{M}}

\newcommand{\bS}{\mathbf{S}}

\newcommand{\bU}{\mathbf{U}}

\newcommand{\bV}{\mathbf{V}}
\newcommand{\bw}{\mathbf{w}}
\newcommand{\bW}{\mathbf{W}}

\newcommand{\rmd}{\mathrm{d}}

\newcommand{\rmF}{\mathrm{F}}

\newcommand{\rmS}{\mathrm{S}}

\newcommand{\bbE}{\mathbb{E}}

\newcommand{\bbN}{\mathbb{N}}

\newcommand{\bbR}{\mathbb{R}}

\newcommand{\bbZ}{\mathbb{Z}}

\DeclareMathAlphabet{\mathbsf}{OT1}{cmss}{bx}{n}
\DeclareMathAlphabet{\mathssf}{OT1}{cmss}{m}{sl}

\DeclareSymbolFont{bsfletters}{OT1}{cmss}{bx}{n}  
\DeclareSymbolFont{ssfletters}{OT1}{cmss}{m}{n}
\DeclareMathSymbol{\bsfGamma}{0}{bsfletters}{'000}
\DeclareMathSymbol{\ssfGamma}{0}{ssfletters}{'000}
\DeclareMathSymbol{\bsfDelta}{0}{bsfletters}{'001}
\DeclareMathSymbol{\ssfDelta}{0}{ssfletters}{'001}
\DeclareMathSymbol{\bsfTheta}{0}{bsfletters}{'002}
\DeclareMathSymbol{\ssfTheta}{0}{ssfletters}{'002}
\DeclareMathSymbol{\bsfLambda}{0}{bsfletters}{'003}
\DeclareMathSymbol{\ssfLambda}{0}{ssfletters}{'003}
\DeclareMathSymbol{\bsfXi}{0}{bsfletters}{'004}
\DeclareMathSymbol{\ssfXi}{0}{ssfletters}{'004}
\DeclareMathSymbol{\bsfPi}{0}{bsfletters}{'005}
\DeclareMathSymbol{\ssfPi}{0}{ssfletters}{'005}
\DeclareMathSymbol{\bsfSigma}{0}{bsfletters}{'006}
\DeclareMathSymbol{\ssfSigma}{0}{ssfletters}{'006}
\DeclareMathSymbol{\bsfUpsilon}{0}{bsfletters}{'007}
\DeclareMathSymbol{\ssfUpsilon}{0}{ssfletters}{'007}
\DeclareMathSymbol{\bsfPhi}{0}{bsfletters}{'010}
\DeclareMathSymbol{\ssfPhi}{0}{ssfletters}{'010}
\DeclareMathSymbol{\bsfPsi}{0}{bsfletters}{'011}
\DeclareMathSymbol{\ssfPsi}{0}{ssfletters}{'011}
\DeclareMathSymbol{\bsfOmega}{0}{bsfletters}{'012}
\DeclareMathSymbol{\ssfOmega}{0}{ssfletters}{'012}

\newcommand{\tilX}{\tilde{X}}

\newcommand{\haty}{\hat{y}}
\newcommand{\hatY}{\hat{Y}}

\newcommand{\tilY}{\tilde{Y}}

\newcommand{\tilZ}{\tilde{Z}}

\newcommand{\bLambda}{\bm{\Lambda}}
\newcommand{\bSigma	}{\bm{\Sigma}}

\DeclareMathOperator*{\argmin}{arg\,min}

\newcommand{\Unif}{\mathrm{Unif}}

\DeclareMathOperator{\tr}{tr}

\DeclareMathOperator{\cov}{\mathsf{Cov}}

\DeclareMathOperator{\rank}{rank}

\usepackage{thmtools, thm-restate}
\newtheorem{theorem}{Theorem} 
\newtheorem{lemma}[theorem]{Lemma}

\newtheorem{proposition}[theorem]{Proposition}

\newtheorem{definition}{Definition} 
\newtheorem{example}{Example} 
 
\newtheorem{remark}{Remark}

\usepackage{cite}

\usepackage{setspace}

\def\BibTeX{{\rm B\kern-.05em{\sc i\kern-.025em b}\kern-.08em
    T\kern-.1667em\lower.7ex\hbox{E}\kern-.125emX}}

\usepackage[utf8]{inputenc} 
\usepackage[T1]{fontenc}    
\usepackage{hyperref}       
\usepackage{url}            
\usepackage{booktabs}       
\usepackage{amsfonts}       
\usepackage{nicefrac}       
\usepackage{microtype}      
\usepackage{xcolor}         
\usepackage{mathtools}
\usepackage{wrapfig}        
\usepackage{cutwin}         
\usepackage{picins} 
\catcode`\@ = 11
\newdimen\@InsertBoxMargin
\newcount\@numlines    
\newcount\@linesleft   
\def\ParShape{%
    \@numlines = 0
    \def\@parshapedata{ }
    \afterassignment\@beginParShape
    \@linesleft
}%
\def\@beginParShape{%
    \ifnum \@linesleft = 0
      \let\@whatnext = \@endParShape
    \else
      \let\@whatnext = \@readnextline
    \fi
    \@whatnext
}%
\def\@endParShape{%
    \global\parshape = \@numlines \@parshapedata
}%
\def\@readnextline#1 #2 #3 {
    \ifnum #1 > 0
      \bgroup  
        \dimen0 = \hsize
        \advance \dimen0 by -#2  
        \advance \dimen0 by -#3  
        \count0 = 0
        \loop
          \global\edef\@parshapedata{%
            \@parshapedata    
            #2                
            \space            
            \the\dimen0       
            \space            
          }%
          \advance \count0 by 1
          \ifnum \count0 < #1
        \repeat
      \egroup
      \advance \@numlines by #1
    \fi
    \advance \@linesleft by -1
    \@beginParShape
}%
\newbox\@boxcontent     
\newcount\@numnormal    
\newdimen\@framewidth   
\newdimen\@wherebottom  
\newif\if@byframe       
\@byframefalse
\def\InsertBoxC#1{%
  \leavevmode
  \vadjust{
    \vskip \@InsertBoxMargin
    \hbox to \hsize{\hss#1\hss}
    \vskip \@InsertBoxMargin
  }%
}%
\def\InsertBoxL#1#2{%
  \@numnormal = #1
  \setbox\@boxcontent = \hbox{#2}%
  \let\@side = 0
  \futurelet \@optionalparameter \@InsertBox
}
\def\InsertBoxR#1#2{%
  \@numnormal = #1
  \setbox\@boxcontent = \hbox{#2}%
  \let\@side = 1
  \futurelet \@optionalparameter \@InsertBox
}%
\def\@InsertBox{%
  \ifx \@optionalparameter [
    \let\@whatnext = \@@InsertBoxCorrection
  \else
    \let\@whatnext = \@@InsertBoxNoCorrection
  \fi
  \@whatnext
}%
\def\@@InsertBoxCorrection[#1]{%
  \ifx \@side 0
    \@@InsertBox{#1}{0}{{\the\@framewidth} 0cm}%
  \else
    \@@InsertBox{#1}{1}{0cm {\the\@framewidth}}%
  \fi
}%
\def\@@InsertBoxNoCorrection{%
  \@@InsertBoxCorrection[0]%
}%
\def\@@InsertBox#1#2#3{%
  \MoveBelowBox
  \@byframetrue
  \@wherebottom = \baselineskip
  \multiply \@wherebottom by \@numnormal
  \advance \@wherebottom by 2\@InsertBoxMargin
  \advance \@wherebottom by \ht\@boxcontent
  \advance \@wherebottom by \pagetotal
  \ifdim \pagetotal = 0cm
    \advance \@wherebottom by -\baselineskip  
  \fi
  \advance \@wherebottom by #1\baselineskip
  \@framewidth = \wd\@boxcontent
  \advance \@framewidth by \@InsertBoxMargin
  \bgroup  
    \ifdim \pagetotal = 0cm
      \dimen0 = \vsize
    \else
      \dimen0 = \pagegoal
    \fi
    \ifdim \@wherebottom > \dimen0
      \immediate\write16{+--------------------------------------------------------------+}%
      \immediate\write16{| The box will not fit in the page. Please, re-edit your text. |}%
      \immediate\write16{+--------------------------------------------------------------+}%
      \vrule width \overfullrule
    \fi
  \egroup
  \prevgraf = 0
  \vbox to 0cm{%
    \dimen0 = \baselineskip
    \multiply \dimen0 by \@numnormal
    \advance \dimen0 by -\baselineskip
    \setbox0 = \hbox{y}%
    \vskip \dp0
    \vskip \dimen0
    \vskip \@InsertBoxMargin
    \ifnum #2 = 1
      \vtop{\noindent \hbox to \hsize{\hss \box\@boxcontent}}%
    \else
      \vtop{\noindent \box\@boxcontent}%
    \fi
    \vss
  }%
  \vglue -\parskip
  \vskip -\baselineskip
  \everypar = {%
    \ifdim \pagetotal < \@wherebottom
      \bgroup  
        \dimen0 = \@wherebottom
        \advance \dimen0 by -\pagetotal
        \divide \dimen0 by \baselineskip
        \count1 = \dimen0
        \advance \count1 by 1
        \advance \count1 by -\@numnormal
        \ifnum #2 = 1
          \ParShape = 3
                      {\the\@numnormal}   0cm   0cm
                      {\the\count1}       0cm   {\the\@framewidth}
                      1                   0cm   0cm
        \else
          \ParShape = 3
                      {\the\@numnormal}   0cm                  0cm
                      {\the\count1}       {\the\@framewidth}   0cm
                      1                   0cm                  0cm
        \fi
      \egroup
    \else
      \@restore@    
    \fi
  }%
  \def\par{%
      \endgraf
      \global\advance \@numnormal by -\prevgraf
      \ifnum \@numnormal < 0
        \global\@numnormal = 0
      \fi
      \prevgraf = 0
  }%
}%
\def\MoveBelowBox{%
  \par
  \if@byframe
    \global\advance \@wherebottom by -\pagetotal
    \ifdim \@wherebottom > 0cm
      \vskip \@wherebottom
    \fi
    \@restore@
  \fi
}%
\def\@restore@{%
    \global\@wherebottom = 0cm
    \global\@byframefalse
    \global\everypar = {}%
    \global\let \par = \endgraf
    \global\parshape = 1 0cm \hsize
}%
\ifx \documentclass \@Dont@Know@What@It@Is@
\else
  \let \pageno = \c@page
\fi

\catcode`\@ = 12

\usepackage{enumitem}       
\usepackage[capitalise]{cleveref}
\usepackage{autonum}        
\usepackage{subcaption} 
\usepackage{makecell} 
\usepackage{cuted}

\newcommand{\KL}{\mathsf{KL}}
\newcommand{\TV}{\mathsf{TV}}
\newcommand{\diam}{\mathsf{diam}}
\newcommand{\qfunc}{\mathsf{Q}}
\newcommand{\gen}{\mathsf{gen}}
\newcommand{\UB}{\mathsf{UB}}
\newcommand{\activ}{\phi}
\newcommand{\tloss}{\tilde{\ell}}
\newcommand{\prodW}{\bW_{\otimes l}}
\newcommand{\prodw}{\bw_{\otimes l}}
\newcommand{\prodWL}{\bW_{\otimes L}}
\newcommand{\prodwL}{\bw_{\otimes L}}

\mathchardef\mhyphen="2D

\newcommand{\cL}{\mathcal{L}}

\newcommand{\cO}{\mathcal{O}}
\newcommand{\cP}{\mathcal{P}}

\newcommand{\cX}{\mathcal{X}}
\newcommand{\cY}{\mathcal{Y}}

\newcommand{\sD}{\mathsf{D}}
\newcommand{\sE}{\mathsf{E}}

\newcommand{\sH}{\mathsf{H}}

\newcommand{\sI}{\mathsf{I}}

\newcommand{\sP}{\mathsf{P}}

\newcommand{\sW}{\mathsf{W}}

\newcommand{\sZ}{\mathsf{Z}}

\newcommand{\NN}{\mathbb{N}}

\newcommand{\RR}{\mathbb{R}}

\title{Information-Theoretic Generalization Bounds for Deep Neural Networks}

\author{Haiyun~He,~\IEEEmembership{Member,~IEEE,} and~Ziv~Goldfeld,~\IEEEmembership{Member,~IEEE}
\thanks{H.\ He is supported by Cornell CAM Postdoctoral Fellowship. 
Z.\ Goldfeld is partially supported by NSF grants CCF-2046018, DMS-2210368, and CCF-2308446, and the IBM Academic Award.} 
\thanks{This paper was  presented in part at the InfoCog workshop at Annual Conference on Neural Information Processing Systems 2023 and in part to International Symposium on Information Theory 2024.} 
\thanks{H.\ He is with the Center for Applied Mathematics, and Z.\ Goldfeld is with the School of Electrical and Computer Engineering, Cornell University, Ithaca, NY 14850 USA. (Emails: \{hh743, goldfeld\}@cornell.edu)}}

\begin{document}

\maketitle

\begin{abstract}

Deep neural networks (DNNs) exhibit an exceptional capacity for generalization in practical applications. This work aims to capture the effect and benefits of depth for supervised learning via information-theoretic generalization bounds. We first derive two hierarchical bounds on the generalization error in terms of the Kullback-Leibler (KL) divergence or the 1-Wasserstein distance between the train and test distributions of the network internal representations. The KL divergence bound shrinks as the layer index increases, while the Wasserstein bound implies the existence of a layer that serves as a generalization funnel, which attains a minimal 1-Wasserstein distance. 
Analytic expressions for both bounds are derived under the setting of binary Gaussian classification with linear DNNs. To quantify the contraction of the relevant information measures when moving deeper into the network, we analyze the strong data processing inequality (SDPI) coefficient between consecutive layers of three regularized DNN models: $\mathsf{Dropout}$, $\mathsf{DropConnect}$, and Gaussian noise injection. This enables refining our generalization bounds to capture the contraction as a function of the network architecture parameters. Specializing our results to DNNs with a finite parameter space and the Gibbs algorithm reveals that deeper yet narrower network architectures generalize better in those examples, although how broadly this statement applies remains a question.

\end{abstract}


\section{Introduction}
\label{Sec: intro}
Overparameterized deep neural networks (DNNs) have become the model of choice for high-dimensional and large-scale learning tasks, thanks to their computational tractability and remarkable generalization performance. Significant efforts have been made to theoretically explain their ability to generalize. These include approaches based on norm-based complexity measures \cite{golowich2018size, neyshabur2017exploring,liang2019fisher}, PAC-Bayes bounds \cite{arora2018stronger,dziugaite2017computing,mcallester1998some,mcallester1999pac,neyshabur2018pac,zhou2018non}, sharpness and flatness of the loss minima \cite{hochreiter1997flat, dinh2017sharp, keskar2016large}, loss landscape \cite{wu2017towards}, and implicit regularization induced by gradient descent algorithms \cite{soudry2018implicit,smith2018bayesian, chatterjee2022generalization}, among others. For a comprehensive literature review, see the recent survey \cite{jakubovitz2019generalization}. Despite this extensive body of research, the precise factors driving the generalization capacity of DNNs remain elusive \cite{zhang2021understanding,kawaguchi2022generalization}. The goal of this work is to shed new light on the advantages of deep models for learning within the framework of information-theoretic generalization bounds.

The generalization error is defined as the difference between the population risk and the empirical risk on the training data. It measures the extent to which a trained neural network whose empirical risk has been pushed to zero overfits the data. Information-theoretic generalization bounds have been widely explored in recent years. This line of work was initiated by \cite{xu2017information,russo2019much}, where a generalization error bound in terms of the mutual information between the input and output of the learning algorithm was derived (see also \cite{bu2020tightening}). These inaugural results inspired various extensions and refinements based on chaining arguments \cite{asadi2018chaining,clerico2022chained}, conditioning and processing techniques \cite{hafez2020conditioning,haghifam2020sharpened,steinke2020reasoning,harutyunyan2021information}, as well as other information-theoretic quantities \cite{esposito2019generalization,aminian2022learning,aminian2021information,wang2019information}. However, these results were not specialized to DNNs and thus did not capture the effect of depth on generalization performance. Quantifying this effect within information-theoretic bounds is the main objective of this work.

\subsection{Main Contributions}

We introduce two novel hierarchical generalization error bounds for deep neural networks (DNNs). The first refines prior results from \cite{xu2017information,russo2019much,bu2020tightening} for sub-Gaussian loss functions, bounding the generalization error through the sum of Kullback-Leibler (KL) divergence and mutual information terms associated with the internal representations at each layer. The mutual information term quantifies the stability required for generalization, ensuring that the algorithm’s output does not depend too strongly on the internal representations of the data. The KL divergence term, on the other hand, contrasts the empirical distribution of the internal representations with their population counterpart, capturing the idea that generalization involves learning the underlying distribution in addition to maintaining output stability. The overall bound shrinks as the layer count increases, underscoring the benefit of deeper architecture for controlling generalization error. Our second generalization bound accounts for Lipschitz continuous losses and employs the 1-Wasserstein distance. This bound suggests the existence of a DNN layer that minimizes the generalization upper bound, acting as a \emph{generalization funnel layer}. The bounds are evaluated for the task of binary Gaussian mixture classification and 
the generalization funnel layer is computed on a simple numerical example, illustrating that it can vary between problems and generally depends on the employed training method.

The hierarchical KL divergence bound qualitatively demonstrates that generalization bounds corresponding to deeper layers are tighter. To quantify the contraction of the bound across layers, we employ the strong data processing inequality (SDPI), which requires requires some stochasticity in the DNN (otherwise, the SDPI coefficient degenerates to 1). We consider three popular randomized regularization techniques: $\mathsf{Dropout}$ \cite{JMLR:v15:srivastava14a}, $\mathsf{DropConnect}$ \cite{wan2013regularization}, and Gaussian noise injection \cite{goldfeld2019estimating,goldfeld2020convergence,gitman2019understanding,neelakantan2015adding,bishop1995training}. 
The SDPI coefficient associated with the stochastic channel induced by each layer is then controlled in terms of network parameters, such as depth, width, activation functions, dropout/noise statistics, etc. The desired generalization bound is then obtained by peeling off the DNN layers and aggregating the corresponding contraction coefficients. 

Our analysis demonstrates that the product of the contraction coefficients across the layers vanishes as the network depth and dropout probabilities (or noise level) increase, or the layer widths decrease. This highlights the advantage of deep network architectures and stochasticity. We also instantiate our results for the Gibbs algorithm \cite{raginsky2017non,aminian2021exact}, yielding an $O(\frac{1}{n})$ generalization bound that decreases monotonically as the product of the contraction coefficients shrinks. Our bounds and their dependence on the problem parameters are visualized via a simple numerical example of a DNN with a finite parameter space. In this instance, we numerically show 
that a deeper but narrower neural network architecture yields a better generalization performance. Overall, our results provide a information-theoretic perspective for understanding the generalization capabilities of DNNs.


\subsection{Related Works}

The generalization performance of deep learning has been extensively studied over the past decades, although a comprehensive and satisfactory account is still obscure. 
Motivated by empirical evidence that overparameterized DNNs tend to generalize better \cite{neyshabur2014search,he2016deep, zhang2021understanding}, numerous attempts to pin this observation down theoretically were made. 
Some control the generalization error using norm-based complexity measures \cite{golowich2018size, neyshabur2017exploring,liang2019fisher}, such as the Rademacher complexity with norm constraints \cite{golowich2018size}, the Frobenius/spectral/$\ell_1$-norm of weight matrices \cite{neyshabur2017exploring}, or the Fisher-Rao norm \cite{liang2019fisher}. A comparisons between several norm-based complexity measures can be found in \cite{neyshabur2017exploring,liang2019fisher}. 
Other works employ PAC-Bayes bounds \cite{mcallester1998some,mcallester1999pac} to understand DNNs in terms of compressibility and robustness \cite{arora2018stronger,dziugaite2017computing,neyshabur2018pac,zhou2018non}. The relationship between generalization and the sharpness/flatness of the loss minima or the landscape of loss was studied in \cite{hochreiter1997flat, dinh2017sharp, keskar2016large, wu2017towards}. The seminal works \cite{zhang2017understanding,zhang2021understanding} analyzed the finite-sample expressivity of DNNs and proved that any two-layer NN with ReLU activation can generalize, given sufficiently many parameters. A numerically-tight DNN generalization bound based on validation and training datasets were derived in \cite{kawaguchi2022generalization}.
  They also suggested that  factors like explicit regularization, minimum norm solution, low complexity, stability, flat minima, etc.,
  many not be necessary for achieving good generalization (see also \cite{zhang2021understanding}). The effect of implicit regularization induced by the gradient descent algorithms on generalization was explored in \cite{soudry2018implicit,smith2018bayesian, chatterjee2022generalization, vardi2023implicit}. 

The information-theoretic study of generalization was initiated by \cite{xu2017information,russo2019much,bu2020tightening}, where the generalization error was controlled by the mutual information between the input and output of the learning algorithm. These results capture the intuition that a learning algorithm generalizes when it extracts relevant information from the training data while disregarding irrelevant information, such as sample noise. This intuition aligns with the compression/clustering phenomena observed between input features and learned representations (cf. \cite{DNNs_Tishby2017,DNNs_ICLR2018,amjad2019learning,goldfeld2019estimating,goldfeld2020information} for related work on the information bottleneck principle for deep learning). Inspired by this novel perspective, many followup works set to derive tighter bounds, e.g., using chaining arguments \cite{asadi2020chaining} or conditional mutual information \cite{hafez2020conditioning,haghifam2020sharpened,steinke2020reasoning}. Information-theoretic bounds that employ other information/discrepancy measures were also explored, encompassing $f$-divergences, $\alpha$-R\'enyi divergences, the generalized Jensen-Shannon divergence, the Wasserstein distance, and more~\cite{esposito2019generalization,aminian2022learning,aminian2021information,lopez2018generalization,wang2019information}. Some works have attempted to obtain information-theoretic generalization bounds by considering specific optimization algorithms. Generalization of the stochastic gradient Langevin dynamics (SGLD) was studied in \cite{pensia2018generalization,haghifam2020sharpened,wang2023generalization}, capturing the effect of training iterations, step size, and noise level. By relating the dynamics in SGLD to stochastic gradient descent (SGD), generalization bounds for SGD were also derived \cite{neu2021information,wang2022on} . 
The most relevant work by Asadi and Abbe \cite{asadi2020chaining} exploited the multilevel structure of DNNs to propose a novel training algorithm based on multilevel entropic regularization.
However, these previous works did not establish generalization bounds that elucidate the the relationship between the strong generalization capability of DNNs and their architecture.
Our goal is to fill this gap by quantifying 
the effects of the network depth, layer width, activation functions, and injected stochasticity on generalization.

\subsection{Paper Outline}
The rest of our paper is organized as follows. In Section II, we define the notations and formulate the supervised learning problem under a feedforward DNN model.  In Section III, we present the hierarchical generalization bounds based on the KL divergence and the Wasserstein distance, respectively. We then specialize our bounds to the case of binary Gaussian classification and derive the analytical expressions. In Section IV, we quantify the contraction of the relevant information measures in the hierarchical KL divergence bounds as the layer index grows. Subsequently, we derive tighter generalization bounds and visualize them using an instance of DNNs with a finite parameter space. Furthermore, we obtain the analytical expressions of the bounds for the Gibbs algorithm. Section V concludes our discussion and outlines avenues for future work. The proofs of our results are provided in the appendices.

\section{Preliminaries and Problem Formulation}
\label{Sec: setup}

\subsection{Notation}
The class of Borel probability measures on $\cX\subseteq \RR^d$ is denoted by $\cP(\cX)$. A random variable $X\sim P_X\in\cP(\cX)$ is called $\sigma$-sub-Gaussian, if $\bbE\big[\exp\big(\lambda(X-\bbE[X])\big)\big]\leq \lambda^2\sigma^2/2$ for any $\lambda\in\bbR$. For two measures $\mu,\nu \in \cP(\cX)$, $\mu\ll\nu$ means that $\nu(E)=0$ implies $\mu(E)=0$ for all measurable set $E$.
The $f$-divergence between $\mu,\nu\in\cP(\cX)$ ($\mu\ll\nu$) is defined by $\sD_{f}(\mu\|\nu)\coloneqq \int f( d  \mu/ d  \nu) \,  d \nu$, where $f:(0,+\infty)\to\bbR$ is convex and $f(1)=0$. The Kullback-Leibler (KL) divergence is defined by taking $f(u)=u\log u$. The Hellinger ($\sH^2$) distance is defined by taking $f(u)=(1-\sqrt{u})^2$. The total variation ($\TV$) distance is defined by taking $f(u)=\frac{1}{2}|u-1|$. 
The mutual information between $(X,Y)\sim P_{X,Y}\in\cP(\cX\times\cY)$ is defined as $\sI(X;Y)\coloneqq \sD_{\mathsf{KL}}(P_{X,Y}\|P_X\otimes P_Y)$. The Shannon entropy of a discrete random variable $X\sim P_X \in \cP(\cX)$ is $\sH(X)=\sH(P_X)=\log(|\cX|)-\sD_\KL(P_X\| \Unif(\cX))$.  For $p\in[1,\infty)$ and a pair of probability measures $\mu,\nu$ on a metric space $(\calX,\|\cdot\|)$ with $\|\cdot\|$ being the Euclidean distance,  the $p$-Wasserstein distance between them is defined as $\sW_p(\mu,\nu)\coloneqq(\inf_{\pi\in\Pi(\mu,\nu)} \bbE_{(x,x')\sim\pi}[\| x-x'\|^p] )^{1/p}$, where $\Pi(\mu,\nu)$ is the set of couplings on $\calX^2$ with marginal distributions $\mu$ and $\nu$. 
For a $d$-dimensional vector $X$ and integers $1\leq i<j\leq d$, we use the shorthands $X_i^j\coloneqq (X_i,\ldots,X_j)$ and  $[j]\coloneqq \{1,2,\ldots,j\}$. For a function $f:\bbR^d\to \bbR^{d'}$, where $f=(f_1,\ldots,f_{d'})$, we define  $\|f\|_\infty \coloneqq \sup_{x\in\RR^{d}}\max_{i=1,\ldots,d'}|f_i(x)|$. For a vector $v$, define $\|v\| \coloneqq \sqrt{v^\intercal v}$ as the Euclidean norm. For a matrix $\bA$, define $\|\bA\|_{\rm{op}}=\sup\{\|\bA v\| \mid \|v\|=1\}$ as the operator norm and $\|\bA\|_\rmF \coloneqq \sqrt{\tr(\bA \bA^*)}$ as the Frobenius norm. 


\subsection{Supervised Learning Problem}
\begin{figure}[t]
    \centering
    \includegraphics[width=.47\textwidth]{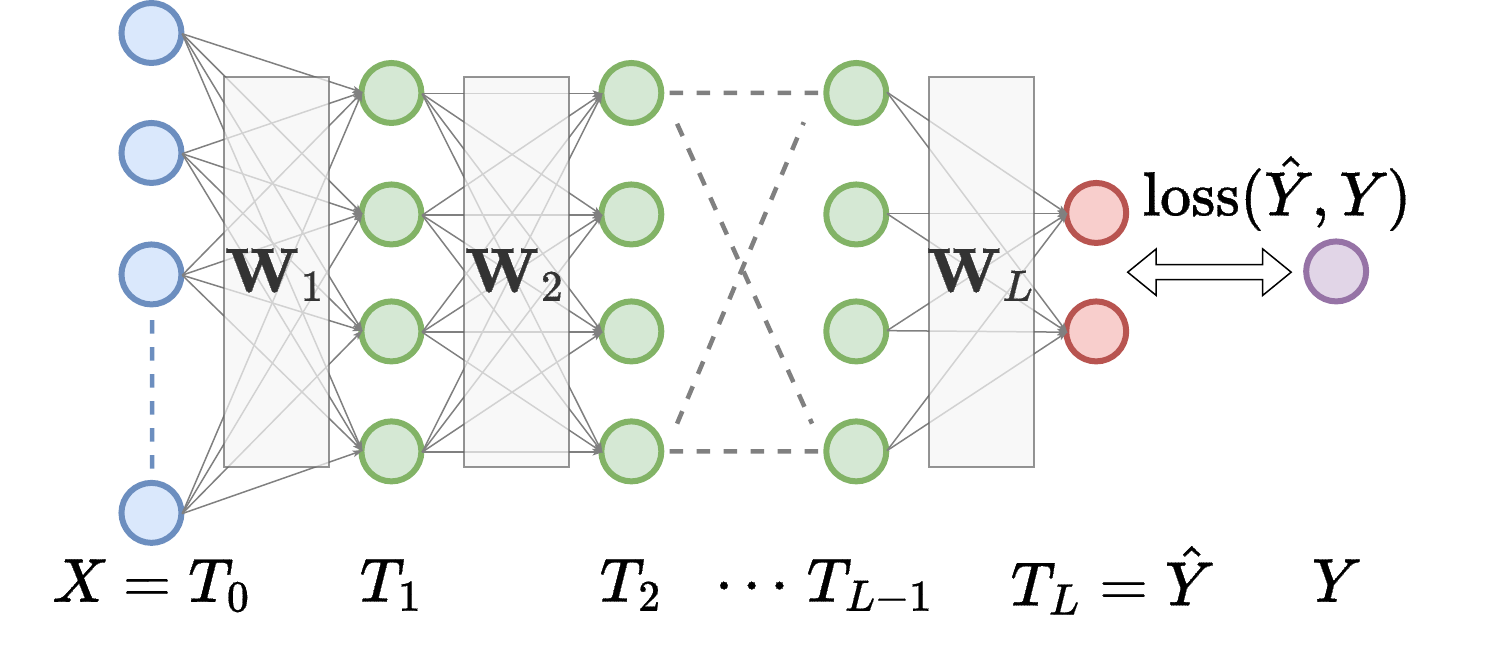}
    \caption{$L$-layer feedforward network.}
    \label{fig:NN diagram}
\end{figure}

Consider a data space $\calX\subseteq \bbR^{d_0}$ and label set $\calY=[K] \subseteq \bbN$. Fix a data distribution $P_{X,Y}\in\calP(\calX \times \calY)$ and let $(X,Y)\sim P_{X,Y}$ be a nominal data feature--label pair. The training dataset $D_n=\{(X_i,Y_i)\}_{i=1}^n$ comprises independently and identically distributed (i.i.d.) copies of $(X,Y)$;  
note that $P_{D_n}=P_{X,Y}^{\otimes n}$. We consider a feedforward DNN model with $L$ layers for predicting the label $Y$ from the test sample $X$ via $\smash{\hat{Y}}\coloneqq g_{\mathbf{w}_L} \circ g_{\mathbf{w}_{L-1}}\circ \cdots \circ g_{\mathbf{w}_1}(X)$, where $g_{\bw_l}(t)=\phi_l(\mathbf{w}_l t)$, $l\in[L]$, for a weight matrix $\mathbf{w}_l \in \bbR^{d_l \times d_{l-1}}$ and an activation function $\phi_l:\mathbb{R}\to\mathbb{R}$ (acting on vectors element-wise). Denote all the network parameters by $\mathbf{w}=(\mathbf{w}_1,\ldots,\mathbf{w}_L)$ and the parameter space by $\calW\subseteq \bbR^{d_1\times d_0}\times\cdots\times\bbR^{d_L\times d_{L-1}}$. 
We denote the internal representation of the $l\textsuperscript{th}$ layer by $T_{l}\coloneqq g_{\mathbf{w}_l} \circ \cdots \circ g_{\mathbf{w}_1}(X)$, $l\in[L]$, noting that $T_{0}=X$. When the input to the network is $X_i$ (rather than $X$), we add a subscript $i$ to the internal representation notation, writing $T_{l,i}$ instead of $T_l$. See Figure \ref{fig:NN diagram} for an illustration. The setup can be generalized to regression problems by setting $\calY\subseteq \bbR$. Furthermore, our arguments extend to the case when the training dataset $D_n$ comprises dependent but identically distributed data samples, e.g., ones generated from a Markov chain Monte Carlo method.

Let $\ell: \calW \times \calX \times \calY  \to \bbR_{\geq 0}$ be the loss function.
Given any $\mathbf{w}\in\calW$, the \emph{population risk} and the \emph{empirical risk} are respectively defined as
\begin{align}
\cL_\sP(\mathbf{w},P_{X,Y})&\coloneqq\bbE[\ell(\bw,X,Y)],\\
    \cL_\sE(\mathbf{w},D_n)&\coloneqq \frac{1}{n}\sum_{i=1}^n \ell(\mathbf{w},X_i,Y_i), 
\end{align}
where the loss function $\ell$ penalizes the discrepancy between the true label $Y$ and the DNN prediction  $\smash{\hat{Y}}= g_{\mathbf{w}_L} \circ \cdots \circ g_{\mathbf{w}_1}(X)$, i.e., $\ell(\bw,x,y)=\tilde\ell(g_{\mathbf{w}_L} \circ \cdots \circ g_{\mathbf{w}_1}(x),y)$. A  learning algorithm trained with $D_n$ can be characterized by a stochastic mapping 
$P_{\mathbf{W}|D_n}$. 
Given any $(P_{\mathbf{W}|D_n},P_{X,Y})$, the \emph{expected generalization error} is defined as the expected gap between the population and empirical risks: 
\begin{equation}
\gen(P_{\mathbf{W}|D_n},P_{X,Y})\coloneqq\bbE[\cL_\sP(\mathbf{W},P_{X,Y})-\cL_\sE(\mathbf{W},D_n)],\label{eq:expected gen-error}
\end{equation}
where the expectation is w.r.t.\ $P_{(X,Y),D_n,\mathbf{W}}=P_{X,Y}^{\otimes (n+1)}P_{\bW|D_n}$.




\section{Hierarchical Generalization Bound} \label{Sec: hierarchical bound}

Existing results such as  \cite{russo2019much,xu2017information,bu2020tightening} bound the generalization error from \eqref{eq:expected gen-error} in terms of the mutual information terms $\sI(D_n;\mathbf{W})$ or $\sum_{i=1}^n \sI(X_i,Y_i;\mathbf{W})$, which only depend on the raw input dataset and the algorithm. We next establish two improved generalization bounds, whose hierarchical structure captures the effect of the internal representations $T_l$. The first bound shrinks as one moves deeper into the network, providing new evidence for the benefits of deep models for learning. The second bound is minimized by one of the network layers, suggesting the existence of a `funnel' layer that governs generalization. 



\subsection{KL Divergence Bound}

We present the following generalization bound for the above described setting.

\begin{theorem}[Hierarchical generalization bound]\label{Thm: NN gen ub}
Suppose that the loss function $\ell(\bw,X,Y)$ is $\sigma$-sub-Gaussian under $P_{X,Y}$, for all $\bw\in\calW$. We have 
\begin{equation}
    \left|\gen(P_{\bW|D_n},P_{X,Y})\right|  \leq \UB(L) \leq \UB(L-1) \leq \ldots \leq \UB(0), \label{Eq: gen ub last layer}
\end{equation}
where
\begin{align}
    &\UB(0)\coloneqq\frac{\sigma\sqrt{2}}{n}\sum\limits_{i=1}^n\!\sqrt{\sI(X_i,Y_i; \bW)}, \nn\\
    &\UB(L)\coloneqq\frac{\sigma\sqrt{2}}{n}\sum\limits_{i=1}^n\!\sqrt{\sD_\KL\big(P_{T_{L,i},Y_i|\bW} \big\| P_{T_{L},Y |\bW}  \big| P_{\bW}\big)}, \quad \text{and}\nn\\
     &\UB(l)\coloneqq\frac{\sigma\sqrt{2}}{n}\sum\limits_{i=1}^n\!\big(\sI(T_{l,i},Y_i; \bW_{l+1}^L| \bW_1^{l}) \\
     &\quad  +\sD_\KL\big(P_{T_{l,i},Y_i|\bW_1^{l}} \big\| P_{T_{l},Y |\bW_1^{l}}  \big| P_{\bW_1^{l}}\big)\big)^{\frac{1}{2}}, l=1,\ldots, L-1. 
\end{align}

\end{theorem}
\cref{Thm: NN gen ub}, proven in \cref{App: pf of Thm: NN gen ub}, is derived by first establishing the $\UB(L)$ upper bound via the Donsker-Varadhan variational representation of the KL divergence and the sub-Gaussianity of the loss function. Conditioning on the parameters of preceding layers, we then invoke the data processing inequality (DPI) to successively peel off layers and derive the remaining bounds. The per-layer bound $\UB(l)$ comprises two terms: the mutual information $\sI(T_{l,i},Y_i; \bW_{l+1}^L| \bW_1^{l})$ quantifies the stability of the learning algorithm in the sense that learned parameters should not depend too strongly on the training data, as represented by the $l$-th layer. The KL divergence term $\sD_\KL\big(P_{T_{l,i},Y_i|\bW_1^{l}} \big\| P_{T_{l},Y |\bW_1^{l}}  \big| P_{\bW_1^{l}}\big)$, on the other hand, captures the other aspect of generalization, whereby the model should learn the underlying data distribution. While the $\UB(L)$ forms the tightest bound, the stated hierarchy highlights the benefit of depth for controlling the generalization error. It mathematically demonstrates that the generalization bound tightens as the number of layers in the neural network increases, allowing for more precise control and a deeper understanding of the behavior of DNNs. Furthermore, the stated hierarchy lends itself well to comparison with existing results. 
Indeed, observing that  $\UB(0)$ is the bound from~\cite{bu2020tightening}, \cref{Thm: NN gen ub} serves as a tightening of that result.
In Section \ref{Sec: noisy network ub}, we present a further discussion about how the depth, width and the stochasticity of DNNs  affect the generalization bounds, which implies that deeper networks have a greater potential for improved generalization compared to shallower ones.

\begin{remark}[Special cases]\label{rem:special_cases}
To gain further intuition, we present two special cases under which the bounds simplify:
\begin{enumerate}[leftmargin=0.5cm]
    \item \underline{One-to-one mapping:} If $g_{\bW_l}$ is one-to-one (i.e., the activation function $\activ_l$ is one-to-one and the weight matrix $\bW_l$ is invertible) for all $l\in[L]$, the DPI holds with equality:
    \begin{align}
        &\sD_\KL\big(P_{T_{l,i},Y_i|\bW_1^{l}} \big\| P_{T_{l},Y |\bW_1^{l}}  | P_{\bW_1^{l}} \big)\\
        &=\sD_\KL\big(P_{X_i,Y_i|\bW_1^{l}} \big\| P_{X,Y }  | P_{\bW_1^{l}} \big)=\sI\big(X_i,Y_i;\bW_1^{l}\big),\\
        \text{and}\quad &\sI\big(T_{l,i},Y_i; \bW_{l+1}^L \big| \bW_1^{l} \big)=\sI\big(X_i,Y_i; \bW_{l+1}^L \big| \bW_1^{l} \big).
    \end{align}
    Thus, the upper bounds are equal: $\UB(L)=\UB(L-1)=\cdots=\UB(0)$,
    which implies that the representation at each layer has the same effect on the generalization error.

    \medskip

  \item \underline{Discrete latent space:} The generalization bound simplifies when $T_l$ can only take finitely many values (e.g., the discrete latent layer in the VQ-VAE \cite{van2017neural}). 
  Assuming that $q_l(\bw_1^l)\coloneqq \min_{t\in\calT_l, y\in\calY}P_{T_{l},Y |\bW_1^{l}}(t,y|\bw_1^l)\in \big(0,|\calT_l\times \calY|^{-1}\big)$,
  we have
  \begin{align} 
      \UB(l)\leq \sqrt{2\sigma^2\log \big(1/ \,\bbE[q_l(\bW_1^l)]\,\big)}. \label{Eq: discrete latent UB} 
  \end{align}
 As $\bbE[q_l(\bW_1^l)]$ grows, we see that $P_{T_{l},Y |\bW_1^{l}}$ tends to the uniform distribution on $T_l\times \calY$ and its entropy/variance increases. This, in turn, shrinks the generalization error, which is consistent with the intuition that stochasticity leads to better generalization.  The proof is provided in Appendix \ref{App: pf for discrete latent}.
  
\end{enumerate}
\end{remark}

\subsection{Wasserstein Distance Bound}
Akin to \cref{Thm: NN gen ub}, we present a generalization error bound based on the Wasserstein distance. Unlike the KL divergence, Wasserstein distances do not generally follow the DPI, and hence the presented bound does not adhere to a descending hierarchical structure. Instead, it shows that there exists a layer that minimizes the Wasserstein generalization bound. 

\begin{theorem}[Min Wasserstein generalization bound]\label{Thm: wasserstein gen ub} 
    Suppose that the loss function $\tloss:\calY\times \calY \to \bbR_{\geq 0}$ is $\rho_0$-Lipschitz and the activation function $\activ_l:\bbR \to \bbR$ is $\rho_l$-Lipschitz, for each $l=1,\ldots,L$. 
    We have
    \begin{align}
        &\gen(P_{\bW|D_n},P_{X,Y}) \!\leq \! \min_{l=0,\ldots, L}\! \frac{\rho_0}{n}\sum_{i=1}^n \bbE_\bW\Bigg[ \!\! \bigg(1\!\vee\!\!\prod_{j=l+1}^L \!\! \rho_j \|\bW_j\|_{\rm{op}} \!\!\bigg)\\
        &\qquad\qquad \qquad \qquad \;  \cdot \sW_1\big(P_{T_{l,i},Y_i|\bW}(\cdot|\bW),P_{T_{l},Y|\bW}(\cdot|\bW) \big) \Bigg].
    \end{align}
\end{theorem}

The derivation of the bound relies on Kantorovich-Rubinstein duality, which ties $\sW_1$ to the difference of expectations defining the generalization error. See \cref{App: pf of wasserstein gen ub} for the proof details. As the Wasserstein distance is monotonically increasing in the order (i.e., $\sW_p\leq \sW_q$ whenever $p\leq q$), the 1-Wasserstein distance provides the sharpest bound. Compared to  \cite[Theorem 2]{wang2019information} and \cite[Theorem 1]{rodriguez2021tighter}, we make a weaker assumption on the loss function (discussed at the end of \cref{App: pf of wasserstein gen ub}) 
by taking Lipschitz continuity of the activation functions into account. 
Compared to the KL divergence bound from \cref{Thm: NN gen ub}, which degenerates when the considered distributions are supported on different domains, the Wasserstein distance is robust to mismatched supports and the corresponding bound is meaningful even in that setting. On the other hand, the 1-Wasserstein bound does not capture the algorithmic stability aspect of generalization, which is quantified by the mutual information term in \cref{Thm: NN gen ub}, and only account for the model learning the underlying population. This may be a consequence of the Wasserstein distance not being an information divergence.

\cref{Thm: wasserstein gen ub} suggests that the generalization bound is controlled by a certain layer that achieves the smallest weighted 1-Wasserstein distance between the distributions of the training and test internal representations. This layer serves as a funnel that determines the overall generalization performance; thus, we call it \emph{generalization funnel layer}. It suggests that within a DNN, there exists a specific layer that exerts a stronger impact on generalization compared to others. In  \cref{Sec: bGMM case} \cref{Ex: eg for Gaussian W bd}, we numerically demonstrate that the generalization funnel layer varies depending on the training method. 

In particular, if $P_{\bW|D_n}$ is symmetric with respect to permutations over the dataset, then $P_{T_{l,i},Y_i|\bW}$ is the same for all $i\in[n]$ and the averaging operation $\frac{1}{n}\sum_{i=1}^n$ can be removed. If $P_{\bW|D_n}$ is deterministic, the bound still depends on the training data distribution and the Wasserstein distance between the distributions of training and test internal representations, which is not vacuous.



\begin{remark}[Comparison with KL divergence bound]\label{Rmk: W comp with KL bd}
Assume that the loss function (bounded within $[0,A]\subset \bbR_{\geq 0}$) and the activation functions in the DNN model satisfy the Lipschitz continuity conditions in Theorem \ref{Thm: wasserstein gen ub}. Under this assumption, the loss function $\ell(\bw,X,Y)$, where $(X,Y)\sim P_{X,Y}$, is $\frac A2$-subGaussian for all $\bw$.
When $\rho_0 K^2 \leq A$, the generalization bound given in Theorem \ref{Thm: wasserstein gen ub} is tighter than $\UB(L)$ in Theorem \ref{Thm: NN gen ub}. A proof of this claim is provided in Appendix \ref{App: pf of Rmk: W comp with KL bd}, and utilizes \cite[Theorem 4]{gibbs2002choosing}, Pinsker's and Bretagnolle-Huber inequalities.
   

\end{remark}

\subsection{Case Study: Binary Gaussian Mixture Classification}\label{Sec: bGMM case}
To better understand the generalization bounds from Theorems \ref{Thm: NN gen ub} and \ref{Thm: wasserstein gen ub} and assess their dependence on depth, we consider the following binary Gaussian mixture example and evaluate the bounds analytically. 

\begin{figure}[t]
    \centering
    \includegraphics[width=.4\textwidth]{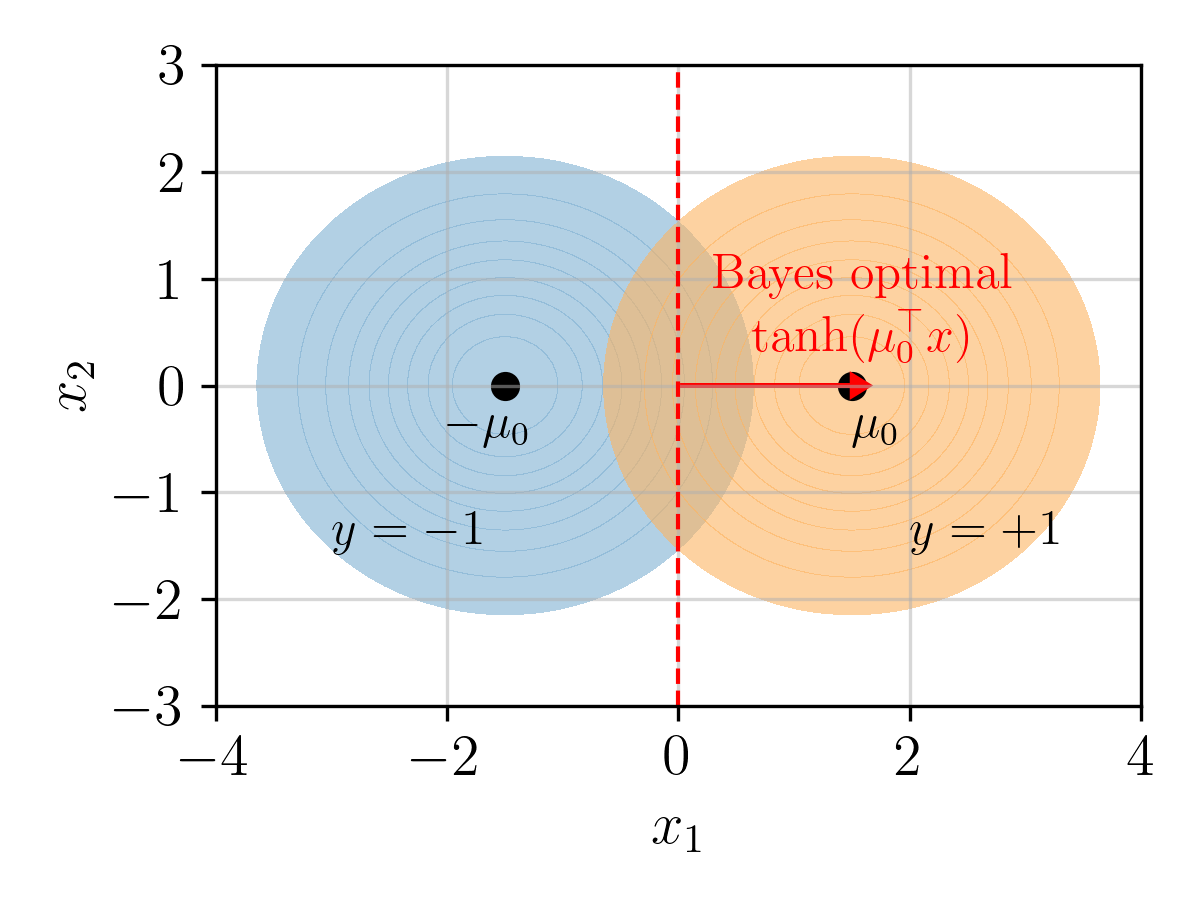}
    \caption{Illustration of binary Gaussian mixture data with $d_0=2$ and the Bayes optimal linear classifier.}
    \label{fig:bGMM}
\end{figure}

\medskip
\noindent\textbf{Classification problem setting.} Consider the binary classification problem illustrated in Fig. \ref{fig:bGMM}, where the input data distribution is a binary Gaussian mixture: $P_Y=\Unif\{-1,+1\}$ and $P_{X|Y=y}=\calN(y\mu_0,\sigma_0^2 \bI_{d_0})$ for $y\in\{\pm 1\}$, %
 where $\mu_0\in \RR^d$ and $\sigma_0>0$. The goal is to classify the binary label $Y$ given the feature $X$. Notice that under this setting, the Bayes optimal classifier is $Y^\star=\tanh(\mu_0^\intercal X)$.

\medskip
\noindent\textbf{Model and algorithm.} Consider a classifier that is realized by a linear $L$-layer neural network composed with a hyperbolic tangent nonlinearity, i.e., $\hatY(\bw)=\tanh(\prodwL X)$, where $\prodw \coloneqq \bw_l\bw_{l-1}\cdots\bw_1$. The output dimension is $d_L=1$ while there are no constraints on the internal dimensions $\{d_l\}_{l=1}^{L-1}$. Therefore, the matrices $\{W_l\}_{l=1}^{L}$ are not required to be invertible. To train the model to approach the Bayes optimal classifier $\tanh(\mu_0^\intercal X)$, we consider a set of algorithms $P_{\bW|D_n}$ under which the output network parameters satisfy $\prodWL^\intercal=\frac{1}{n}\sum_{i=1}^n Y_iX_i$, and set the prediction to $\hatY=\hatY(\bW)$. The rationale behind this choice of algorithm comes from observing that $Y_iX_i \sim \calN(\mu_0,\sigma_0^2\bI_{d_0})$ are i.i.d. for $i=1,\ldots,n$, and by the strong law of large numbers we have $\prodWL^\intercal\to \mu_0$ almost surely, as $n\to\infty$. Performance is measured using the quadratic loss function $\ell(\bw,X,Y)=\big(Y-\tanh(\bw_{\otimes L} X)\big)^2$, which is bounded inside $[0,4]$ and is thus $2$-sub-Gaussian under $P_{X,Y}$, for all~$\bw$.


\medskip
\noindent\textbf{Analysis.} We move to evaluate the generalization bounds in Theorems \ref{Thm: NN gen ub} and \ref{Thm: wasserstein gen ub} by computing the prior and posterior distributions and the divergences between them. Proofs of subsequent claims are all deferred to Appendix \ref{App: pf of gaussian example KL bound}.

\begin{lemma}[Prior and posterior of $(X_i,Y_i)$]
     For any $i=1,\ldots,n$ and $y\in\{\pm 1\}$, the prior distribution of $X_i|Y_i=y$ is given by $P_{X_i|Y_i=y}=\calN(y\mu_0,\sigma_0^2\bI_{d_0})$, while its posterior distribution given a model $\prodWL=\prodwL$ is $P_{X_i|Y_i=y,\prodWL=\prodwL}=\calN(y\prodwL^\intercal,\frac{(n-1)\sigma_0^2}{n}\bI_{d_0} )$. We also have $P_{Y_i|\prodWL=\prodwL}=P_{Y_i}=\Unif\{-1,+1\}$. 
\end{lemma}

Given the above expressions for the involved distributions, we evaluate the KL divergence generalization bound from \cref{Thm: NN gen ub} as follows.

\begin{proposition}[KL divergence bound evaluation]\label{Prop: Gaussian KL bd}
Under the binary Gaussian classification setting, we have
    \begin{align}
        \left|\gen(P_{\bW|D_n},P_{X,Y})\right|\leq \widetilde{\UB}(L)\leq \widetilde{\UB}(L-1)\leq \cdots \leq \widetilde{\UB}(0),
    \end{align}
    where $\widetilde{\UB}(l)\coloneqq 2\sqrt{\bbE_\bW[r_l] (\log \frac{n}{n-1}-\frac{1}{n})+\frac{d_0}{n}}$, $r_0=d_0$, and $r_l=\rank(\prodW)$, for $l\in[L]$.
\end{proposition}
As a sanity check, observe that $\widetilde{\UB}_n(l)$ converges to $0$ as $n\to\infty$, for all $l=0,1,\ldots,L$, as expected. Recalling that $\rank(\bA\bB)\leq \rank(\bA)\wedge \rank(\bB)$, we see that $r_L\leq r_{L-1} \leq \cdots\leq r_1 \leq r_0=d_0$. Consequently, the contraction from $\widetilde{\UB}_n(l-1)$ to $\widetilde{\UB}_n(l)$ is evident and quantified by the gap between the ranks of $\bW_{\otimes(l-1)}$ and $\prodW$, namely, $r_{l-1}-r_l$. Note that in our example, $d_L=1$, $1\leq \rank(\bW_L)\leq d_L \wedge d_{L-1}$ and thus $\rank(\prodWL)=\rank(\bW_L)=1$, independent of the depth $L$, which means that the tightest bound, $\widetilde{\UB}(L)$, does not change with $L$. Nevertheless, the intermediate bounds $\widetilde{\UB}(l)$, for $l\in[L-1]$, generally shrink as $L$ grows, depicting the trajectory of generalization performance of the internal layers.  Extending the above example beyond the classification setting to representation learning, where the output representation dimension $d_L$ varies according to the network structure, would enable observing a similar effect for $\widetilde{\UB}(L)$ as well. Our focus on the binary classification case is motivated by its analytic tractability, and we leave further extensions for future work. 




\medskip

We proceed to evaluate the Wasserstein generalization bound under the considered setting. 

\begin{proposition}[Wasserstein distance based bound evaluation]\label{Prop: Gaussian W bd}
Under the binary Gaussian classification setting and from \cref{Thm: wasserstein gen ub}, we have
\begin{align}
        &\gen(P_{\bW|D_n},P_{X,Y}) \leq  \frac{4\sqrt{2}\sigma_0(\sqrt{d_0}+(\sqrt{n}-\sqrt{n-1})) }{\sqrt{n}} \\
        &\cdot \min_{l=0,\ldots, L} \bbE_\bW\left[ \bigg(1\vee \prod_{j=l+1}^L \|\bW_j\|_{\rm{op}}^2 \bigg)\|\prodW \|_\rmF^2 \right]^{\frac{1}{2}} ,
\end{align}
where $\prodW= \bW_l \bW_{l-1}\cdots \bW_1$ for $l\in[L]$, and $\bW_{\otimes 0}=\bI_{d_0}$.
\end{proposition}
This upper bound is computed using the 2-Wasserstein distance between two Gaussian distributions based on the fact that $\sW_1\leq \sW_2$. Note that this upper bound also vanishes as $n\to\infty$. Specifically,  both bounds in Propositions \ref{Prop: Gaussian KL bd} and \ref{Prop: Gaussian W bd} are on the same order of $O(\frac{1}{\sqrt{n}})$.
In this case, the \emph{generalization funnel} layer that yields the tightest upper bound depends on the Frobenius norm of the product of network weight matrices up to the current layer $\|\prodW\|_\rmF$ and the product of subsequent layers' operator norms. We notice that $\|\prodW\|_\rmF=\sqrt{\tr(\prodW\prodW^\intercal)}$ not only depends on $\rank(\prodW)$ but also on the singular values of $\prodW$. Thus,  the generalization funnel layer is not necessarily the last one. Additionally, further minimizing this upper bound over the learning algorithm $P_{\bW|D_n}$ as well as the DNN’s depth and width would be interesting, though highly non-trivial. We leave this for future investigation.
In the following example, by considering a simple neural network model with different training methods, we  empirically show that the generalization funnel layer depends on the training method. It implies that there indeed exists a network layer that plays a more important role in controlling the generalization performance of the learning algorithm.

\begin{example}[Numerical evaluation of \cref{Prop: Gaussian W bd}]\label{Ex: eg for Gaussian W bd}

Consider a DNN with $L=10$ layers, each of width 2, i.e., $d_0=d_1=\cdots=d_{L-1}=2$. Let the training dataset $D_n$ contain $n=100$ samples that are i.i.d. from the aforementioned binary Gaussian mixture with $\mu_0=(0.5\ 0)^\intercal$ and $\sigma_0=1$. We consider a learning algorithm $P_{\bW|D_n}$ that proceeds as follows: randomly generate $(2\times 2)$ rotation matrices $\bW_1,\ldots, \bW_{L-1}$ and the vector $\bW_L$ such that $\prodWL^\intercal = \frac{1}{n}\sum_{i=1}^n Y_iX_i$. Specifically, let $\bW_l=C_l\big(\begin{smallmatrix}
  \cos{\theta_l} & \sin{\theta_l}\\
  -\sin{\theta_l} & \cos{\theta_l}
\end{smallmatrix}\big)$, for $l=1,\ldots,L-1$, be the rotation matrix, and set the last layer parameter vector as  $\bW_L=(0,C_L)$. The rotation angles $\theta_l$ and scaling factors $C_l$ are then calibrated based on $D_n$ to yield the desired distribution, i.e., $\prodWL = \frac{1}{n}\sum_{i=1}^n Y_iX_i$. To that end, we let $\theta_l=\frac{1}{L-1}\arccos{\langle \bW_L, \frac{1}{n}\sum_{i=1}Y_iX_i \rangle}$ for $l=1,\ldots, L-1$, while randomly generate $\{C_l\}_{l=1}^L$ such that $\prod_{l=1}^{L}C_l=\|\frac{1}{n}\sum_{i=1}Y_iX_i\|$. 


Under this algorithm, we have that $\prodW$ is full-rank, $\|\prodW\|_\rmF=\sqrt{2}\prod_{j=1}^l C_j$ for $l=1,\ldots,L-1$, $\|\prodWL\|_\rmF=\prod_{l=1}^{L}C_l=\|\frac{1}{n}\sum_{i=1}Y_iX_i\|$ and $\|\bW_l\|_{\rm{op}}=C_l$ for $l=1,\ldots,L$. Given any training dataset $D_n$, the generalization funnel layer is determined by the way we generate $\{C_l\}_{l=1}^L$. We first let $\prod_{j=l+1}^L C_j\leq 1$ for all $l=0,1,\ldots,L-1$. If we pick any $1\leq l'\leq L-1$ and let $\prod_{j=1}^{l'}C_j$ be sufficiently small (at least smaller than $\sqrt{d_0} \wedge \|\frac{1}{n}\sum_{i=1}Y_iX_i\|$),  then the generalization funnel layer is equal to $l'$. For numerical illustration, we generate $L$ numbers from the uniform distribution $\Unif\{[0,1]\}$ and scale them to be $C_l$'s such that $\prod_{l=1}^L C_l = \|\frac{1}{n}\sum_{i=1}^n Y_i X_i\|$ and $\prod_{j=1}^{l'}C_j=0.2\|\frac{1}{n}\sum_{i=1}Y_iX_i\|$, for $l'\in\{3,5,7\}$. 
We compute the generalization funnel layer index as the minimizer of the sample mean from $10^4$ output network parameters $\bW$ (trained on $100$ datasets $D_n)$: $l^*=\argmin_{l\in[0:L]} \text{SampleMean}\big((1 \vee \prod_{j=l+1}^L \|\bW_j\|^2)\|\prodW\|_\rmF^2  \big) = \argmin_{l\in[0:L]} \text{SampleMean}\big(\|\prodW\|_\rmF^2  \big)$.
As shown in Table \ref{tab:W bd eval},  the generalization funnel layer varies according to the parameter generating methods.
\begin{table}[t] 
    \begin{center}
    \caption{The generalization funnel layer index $l^*$ for differently generated model $\bW$ when $L=10$  in Example \ref{Ex: eg for Gaussian W bd}. }
    \label{tab:W bd eval}
    \begin{tabular}{|c|c|c|c|}
       \hline
       \makecell{Generating methods \\ $\prod_{j=1}^{l'}C_j=0.2\|\frac{1}{n}\sum_{i=1}Y_iX_i\|$} & $l'=3$ & $l'=5$ & $l'=7$\\ \hline
       Generalization funnel layer $l^*$ & 3 & 5 & 7\\
       \hline
    \end{tabular}
    \end{center}
\end{table}
\end{example}

\section{Tighter Generliazation Bound via Contraction}\label{Sec: noisy network ub}

Inspired by Theorem \ref{Thm: NN gen ub}, we next aim to capture the contraction of the information measures in our bound as the layer count grows. To that end, we use the \emph{strong} DPI (SDPI) \cite{COHEN1993211,dobrushin1956central, polyanskiy2017strong}, for which some preliminaries are needed.

\subsection{Strong Data Processing Inequality}
 Given $P_X,Q_X\in\cP(\cX)$ and a transition kernel (channel) $P_{Y|X}$, write $P_Y=P_{Y|X}\circ P_X$ and $Q_Y=P_{Y|X}\circ Q_X$ for the marginal distributions at the output of the channel when we feed it with $P_X$ or $Q_X$, respectively. Assuming $P_X\ll Q_X$ and that $Q_X$ is not a point mass, the SDPI coefficient for $P_{Y|X}$ under the $f$-divergence $\mathsf{D}_f$ is
\begin{align}
    \eta_f(P_{Y|X})&\coloneqq \sup_{P_X, Q_X}\frac{\sD_f(P_{Y|X}\circ P_X \| P_{Y|X}\circ Q_X)}{\sD_f(P_X\|Q_X)} \in [0,1].
\end{align}
We write $\eta_\KL(P_{Y|X})$ and $\eta_\TV(P_{Y|X})$ for the coefficients under the KL divergence and the total variation distance, respectively. Proposition II.4.10 in \cite{cohen1998comparisons} shows that for any $f$-divergence, we have
\begin{align}
    \eta_f(P_{Y|X}) \leq \eta_{\,\TV}(P_{Y|X}) =\sup_{x,x'\in\calX}\|P_{Y|X=x}-P_{Y|X=x'}\|_{\TV},
\end{align}
where $\eta_{\,\TV}(P_{Y|X})$ is also know as \emph{Dobrushin's coefficient} \cite{dobrushin1956central}.  
It can be shown that if $Y=g(X)$ for some deterministic function $g:\calX \to \calY$, then $\eta_f(P_{g(X)|X})=\eta_{\,\TV}(P_{g(X)|X})=1$ (cf., e.g., Proposition II.4.12 from \cite{cohen1998comparisons}). 

 According to the above, if all the feature maps $g_{\mathbf{w}_l}$, for $l=1,\ldots,L$, in the DNN are deterministic, the contraction coefficients we are looking for  degenerate to 1, landing us back at the bound from Theorem \ref{Thm: NN gen ub}. To arrive at a nontrivial contraction coefficient, we consider the following common regularization techniques in machine learning:    $\mathsf{Dropout}$ \cite{JMLR:v15:srivastava14a}, $\mathsf{DropConnect}$ \cite{wan2013regularization}, and Gaussian noise injection \cite{gitman2019understanding,neelakantan2015adding,bishop1995training,goldfeld2019estimating}. These methods have been used to introduce stochasticity to neural networks to enhance generalization and robustness.

Before presenting the subsequent results, we find it succinct to define the notations: for any vector $V$, let $V(i)$ be $i\textsuperscript{th}$ element of $V$; for any matrix $\bM$, let $\bM(i,j)$ be the $(i,j)\textsuperscript{th}$ element of $\bM$.

\begin{figure*}[!t]
\begin{subfigure}[t]{.33\textwidth}
\centering
\includegraphics[width=\textwidth]{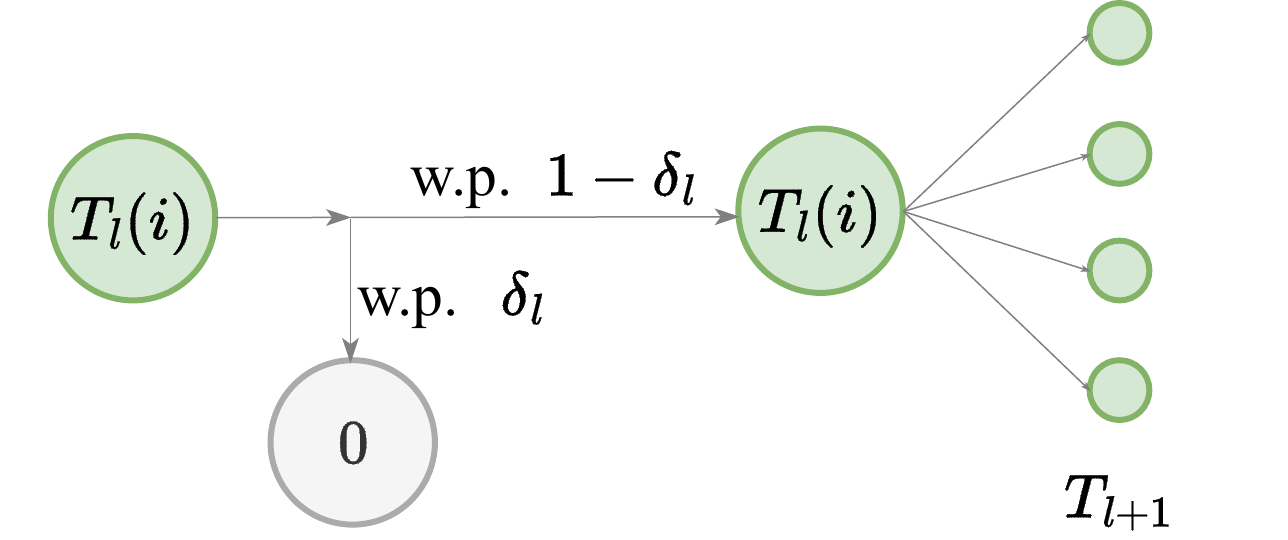}
\caption{}
\label{Fig: dropout}
\end{subfigure}
\begin{subfigure}[t]{0.33\textwidth}
\centering
\includegraphics[width=\textwidth]{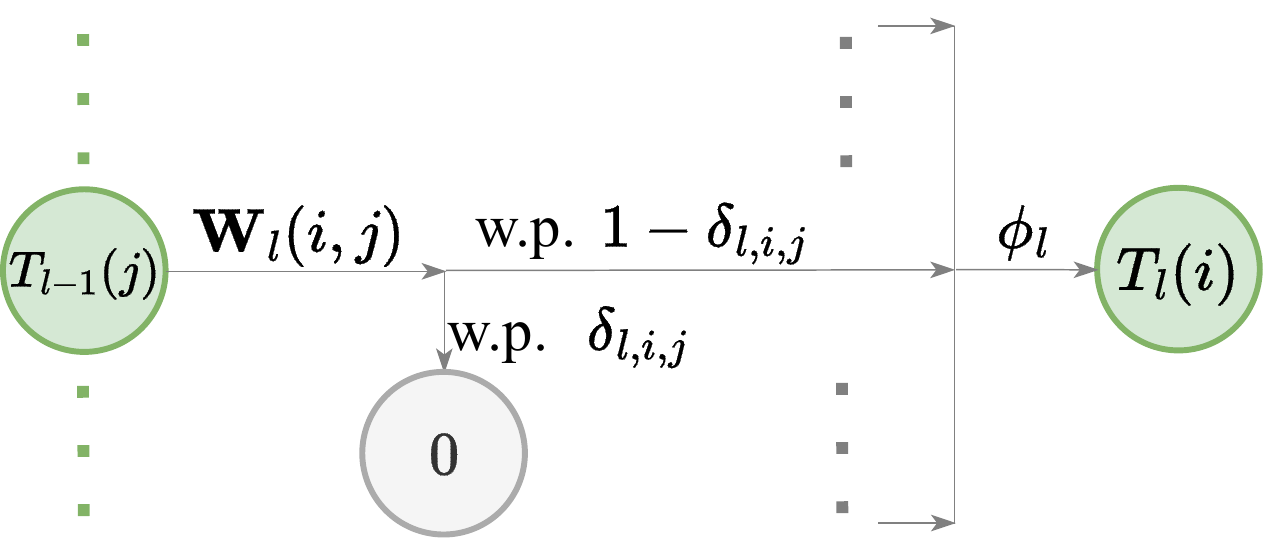}
\caption{}
\label{Fig: dropconnect}
\end{subfigure}
\begin{subfigure}[t]{0.33\textwidth}
\centering
\includegraphics[width=\textwidth]{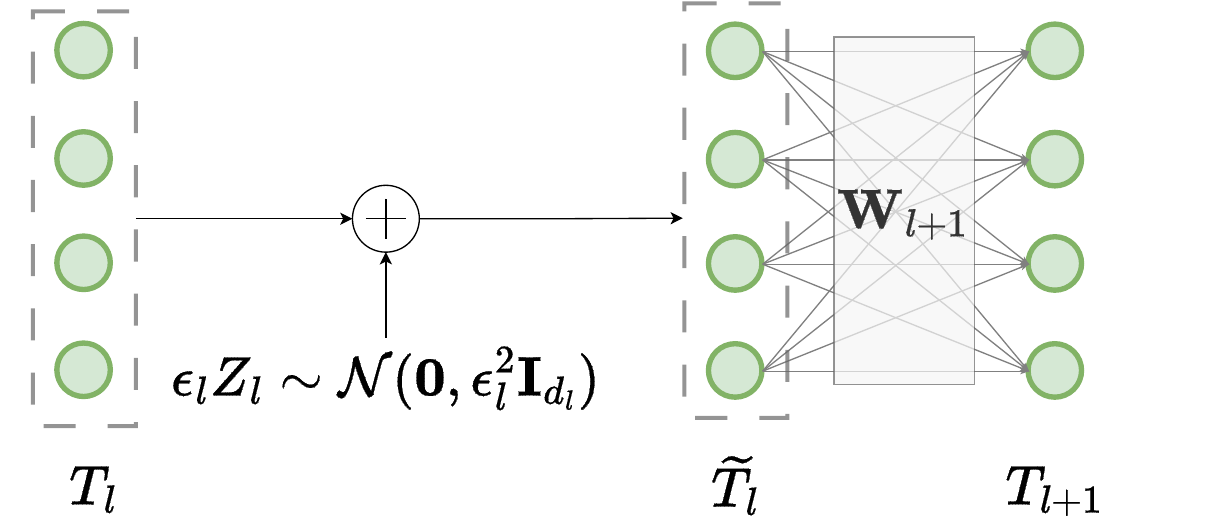}
\caption{}
\label{Fig: gaussian noise NN}
\end{subfigure}
\caption{Examples of DNNs with stochascity. (a) $l\textsuperscript{th}$ layer with $\mathsf{Dropout}$  probability $\delta_l$. (b) $l\textsuperscript{th}$ layer with $\mathsf{DropConnect}$  probabilities $\{\delta_{l,i,j}\}_{j=1}^{d_{l-1}}$. (c) Noisy DNN with injected  isotropic Gaussian noise to the $l\textsuperscript{th}$ layer. }
\end{figure*}

\subsection{Generalization Bound for DNNs with \texorpdfstring{$\mathsf{Dropout}$}{Dropout}}\label{Sec: gen bd for dropout}

The  $\mathsf{Dropout}$ \cite{JMLR:v15:srivastava14a} technique has been shown to be a simple way to prevent neural networks from overfitting. We hereby analyse the contraction behaviour in the generalization bounds caused by  $\mathsf{Dropout}$.
Consider the $l\textsuperscript{th}$ layer of the neural network that applies a $\mathsf{Dropout}$ probability $\delta_l\in(0,1)$, $l=0,\ldots,L-1$, as shown in Figure \ref{Fig: dropout}. Let $E_l\sim \mathsf{Bern}(1-\delta_l)^{\otimes d_l}$ 
be an i.i.d. sequence of $d_l$ Bernoulli variables. Then the activation output of the $(l+1)\textsuperscript{th}$ layer is given by
\begin{align}
    T_{l+1}=\activ_{l+1}(\bW_{l+1}(T_{l} \odot E_{l}))=: \activ_{l+1}(\bW_{l+1} \widetilde{T}_{l}), \label{Eq: dropout def}
\end{align}
where $\odot$ denotes the elementwise product operation. When $T_l(i)$ is not equal to $0$, with probability $\delta_l$, $T_l(i)$ is deactivated as $0$ before being passed to the next layer; with probability $1-\delta_l$, $T_l(i)$ retains its value. When $T_l(i)=0$, it always remains unchanged. The channel $P_{\widetilde{T}_l|T_l}$ can then be regarded as a composition of $d_l$ parallel independent generalized $\mathsf{Z}$-channels \cite{10.5555/1146355}.
Thus, the stochastic channel $P_{T_{l+1}|T_l,\bW}=P_{T_{l+1}|\widetilde{T}_l,\bW}\circ P_{\widetilde{T}_l|T_l}$ yields a non-trivial SDPI coefficient. 

\begin{lemma}[$\mathsf{Dropout}$ SDPI coefficient]\label{Lem: dropout SDPI coeff}
The SDPI coefficient of the $\mathsf{Dropout}$ channel $P_{\widetilde{T}_l|T_l}$ with $\mathsf{Dropout}$ probability $\delta_l \in(0,1)$ is $\eta_\KL(P_{\widetilde{T}_l|T_l})=1-\delta_l^{d_l}$, for $l=0,\ldots,L$.
\end{lemma}
\cref{Lem: dropout SDPI coeff}, which is proven in Appendix \ref{App: pf of dropout}, implies that whenever $\mathsf{Dropout}$ is applied to a certain layer, it leads to a strict contraction in the generalization bound in \cref{Thm: NN gen ub} since the coefficient $\eta_\KL(P_{\widetilde{T}_l|T_l})$ is strictly less than 1. Consequently, the layer mappings in a DNN with $\mathsf{Dropout}$ gives rise to the following result.

\begin{theorem}[DNN with $\mathsf{Dropout}$ generalization bound]\label{Thm: dropout DNN gen ub}
Consider the DNN model with $\mathsf{Dropout}$ probability $\delta_l\in(0,1)$ at the $l\textsuperscript{th}$ layer (cf.\ \eqref{Eq: dropout def}), $l=0,\ldots,L-1$. If the loss function $\ell(\bw,X,Y)$ is $\sigma$-sub-Gaussian under $P_{X,Y}$, for all $\bw\in\calW$, we have
\begin{align}
    &| \gen(P_{\bW|D_n},P_{X,Y}) |\\
    &\leq \frac{\sigma\sqrt{2}}{n}\sum_{i=1}^n\sqrt{\prod_{l=0}^{L-1}(1-\delta_l^{d_l})\sI(X_i;\bW|Y_i)+\sI(Y_i;\bW) }.
\end{align}
\end{theorem}
The proof of \cref{Thm: dropout DNN gen ub} initiates at the bound $\UB(L)$ from Theorem \ref{Thm: NN gen ub} and first factors out the terms that depend on the label $Y$ from the KL divergence. This yields the summand $\sI(Y_i;\bW)$. This step is necessary since the label is not processed by the DNN with $\mathsf{Dropout}$, and the corresponding SDPI coefficient (without factoring $Y_i$ out) would degenerate to 1. For the remaining term, we invoke the SDPI $L$ times collecting the coefficients and invoking Lemma \ref{Lem: dropout SDPI coeff} to arrive at the desired bound. The details are provided in \cref{App: pf dropout gen ub}.

    We make the following observations regarding the bound from \cref{Thm: dropout DNN gen ub}. 
    The coefficient product $\prod_{l=0}^{L-1}(1-\delta_l^{d_l})$ exhibits a decrement from 1 to 0 with: (i) increasing  $\mathsf{Dropout}$ probabilities $\{\delta_l\}_{l=0}^{L-1}$; (ii) elevated network depth $L$; and (iii) shrinking layer widths $\{d_l\}_{l=0}^{L-1}$. In addition, using Taylor's expansion, we have $\log(1-\delta^{d_l})\approx -\delta_l^{d_l}$ and $\delta_l^{d_l}=(1-(1-\delta_l))^{d_l}\approx 1-d_l(1-\delta_l)$. Then the product of coefficient can be approximated as 
    \begin{align}
    \prod_{l=0}^{L-1}(1-\delta_l^{d_l})&= \exp\left(\sum_{l=0}^{L-1}\log(1-\delta_l^{d_l})\right)\approx \exp\left(-\sum_{l=0}^{L-1}\delta_l^{d_l}\right) \nn\\
    &\approx \exp\left(-\sum_{l=0}^{L-1}\big(1-d_l(1-\delta_l)\big)\right), \nn
   \end{align}
highlighting an exponential decay. Since the label follows a categorical distribution of parameter $K$, we have $\sI(Y_i;\bW)\leq \log K$.
The smaller the number of distinct labels $K$, the better the generalization bound. 
The behavior of $\sI(X_i;\bW|Y_i)$, on the other hand, is harder to pin down as it depends on the data distribution and the learning algorithm at hand. In the following statement, we investigate the behavior of $\sI(X_i,Y_i;\bW)$ w.r.t.\ the input layer $\mathsf{Dropout}$ probability $\delta_0$. At the end of \cref{Sec: noisy network ub}, we investigate the generalization  bound w.r.t.\ the layer dimension and network depth with an instance of DNN with a finite parameter space.

\begin{lemma}[Upper bound on $\sI(X_i,Y_i;\bW)$ in DNNs with $\mathsf{Dropout}$]\label{Coro: Dropout bd mono-shrink}
    For a DNN model with $\mathsf{Dropout}$ probability $\delta_0 \in(0,1)$ at the input layer, we have for all $i=1,\ldots,n$,
    \begin{align}
        \sI(X_i,Y_i;\bW)\leq \mathsf{MIUB}_i(\delta_0) \coloneqq \sum_{k=0}^{d_0} \delta_0^{d_0-k}(1-\delta_0)^{k}\binom{d_0}{k} I_k^{(i)},
    \end{align}
    where $I_k^{(i)}=\max_{\calJ(k)\subseteq [d], |\calJ(k)|=k} \sI\big((X_i(j))_{j\in \calJ(k)},Y_i;\bW\big)$, for $k=0,\ldots,d_0$. The upper bound $\mathsf{MIUB}_i(\delta_0)$ monotonically shrinks to $0$ as $\delta_0$ increases from $0$ to $1$.
\end{lemma}
The proof of \cref{Coro: Dropout bd mono-shrink}, fully provided in \cref{App: pf of Coro: Dropout bd mono-shrink}, hinges on the application of the chain rule of mutual information. \cref{Coro: Dropout bd mono-shrink} suggests that the generalization bound (c.f.\ $\UB(0)$ in \cref{Thm: NN gen ub}) shrinks even when $\mathsf{Dropout}$ is only applied to the input data, commonly employed as a data augmentation technique.
Since $\sI(X_i;\bW|Y_i), \sI(Y_i;\bW)\leq \sI(X_i,Y_i;\bW)$, 
\cref{Thm: dropout DNN gen ub} and \cref{Coro: Dropout bd mono-shrink} show that the generalization bound of a DNN with $\mathsf{Dropout}$ diminishes as the $\mathsf{Dropout}$ probabilities increase. This aligns with the intuition that introducing stochasticity, as facilitated by $\mathsf{Dropout}$ mechanisms, enhances the model's capacity for generalization. 

\subsection{Generalization Bound for DNNs with \texorpdfstring{$\mathsf{DropConnect}$}{DropConnect}}\label{Sec: gen bound dropconnect}
Compared to $\mathsf{Dropout}$ which effectively deactivates some columns of the network weight matrices, $\mathsf{DropConnect}$ \cite{wan2013regularization} operates in a more precise way by randomly deactivating each element of the weight matrices. Consider the $l\textsuperscript{th}$ layer of the neural network adopts a $\mathsf{DropConnect}$ probability  $\delta_{l,i,j}\in(0,1)$ at the $(i,j)\textsuperscript{th}$ entry of the weight matrix $\bW_l$, $l=1,\ldots,L$, as shown in Figure \ref{Fig: dropconnect}. Let $\bE_l \sim \prod_{i=1}^{d_l}\prod_{j=1}^{d_{l-1}}\mathsf{Bern}(1-\delta_{l,i,j})$ be a $d_l\times d_{l-1}$ binary matrix with i.i.d.\ Bernoulli elements. 
Then the activation output of the $l\textsuperscript{th}$ layer is given by
\begin{align}
    T_{l}&=\activ_{l}((\bW_{l}\odot \bE_{l})T_{l-1})\\
    &=: \bigg[\activ_{l}\bigg(\sum_{j=1}^{d_{l-1}} \bW_{l}(1,j)\widetilde{T}_{l-1}(1,j)\bigg),\ldots,\\
    &\qquad \; \activ_{l}\bigg(\sum_{j=1}^{d_{l-1}} \bW_{l}(d_l,j)\widetilde{T}_{l-1}(d_l,j)\bigg)\bigg]^\intercal, \label{Eq: dropconnect layer}
\end{align}
where $\odot$ denotes the elementwise product operation, $\widetilde{T}_{l-1}$ is a $d_{l}\times d_{l-1}$ matrix, and $\widetilde{T}_{l-1}(i,j):=\bE_{l}(i,j)T_{l-1}(j)$. When $\bW_{l}(i,j)T_{l-1}(j)$ is not equal to $0$, with probability $\delta_{l,i,j}$, $\bW_{l}(i,j)T_{l-1}(j)$ is deactivated as 0 before being passed to the $i^\textsuperscript{th}$ neuron of the next layer; with probability $1-\delta_{l,i,j}$, $\bW_{l}(i,j)T_{l-1}(j)$ keeps its value. When $\bW_{l}(i,j)T_{l-1}(j)=0$, it always remains unchanged. In a similar manner to \cref{Sec: gen bd for dropout}, the channel $P_{\widetilde{T}_{l-1}|T_{l-1}}$ can then be regarded as a composition of $d_{l}d_{l-1}$ parallel independent generalized $\sZ$-channels. Thus, the stochastic channel $P_{T_{l}|T_{l-1},\bW}=P_{T_{l}|\widetilde{T}_{l-1},\bW}\otimes P_{\widetilde{T}_{l-1}|T_{l-1}}$ also yields a non-trivial SDPI coefficient. 

\begin{lemma}[$\mathsf{DropConnect}$ SDPI coefficient bound]\label{Lem: dropconnect SDPI coeff}
The SDPI coefficient of the $\mathsf{DropConnect}$ channel $P_{\widetilde{T}_{l-1}|T_{l-1}}$ with $\mathsf{DropConnect}$ probabilities $\delta_{l,i,j}\in(0,1)$, $i=1,\ldots, d_l, j=1,\ldots, d_{l-1}$, satisfies $\eta_\KL(P_{\widetilde{T}_{l-1}|T_{l-1}})\leq 1-\prod_{i=1}^{d_l}\prod_{j=1}^{d_{l-1}}\delta_{l,i,j}$, for $l=1,\ldots,L$.
\end{lemma}
\cref{Lem: dropconnect SDPI coeff}, which is proven in Appendix \ref{App: pf of Lem: dropconnect SDPI coeff}, implies that the application of $\mathsf{DropConnect}$ at a certain layer also leads to a strict contraction in the generalization bound in \cref{Thm: NN gen ub} since the coefficient $\eta_\KL(P_{\widetilde{T}_{l-1}|T_{l-1}})<1$. Compared to the coefficient shown in \cref{Lem: dropout SDPI coeff}, this coefficient decreases with an increase in individual probability $\delta_{l,i,j}$. This obsesrvation  highlights the advantageous flexibility of $\mathsf{DropConnect}$ in improving generalization performance.
Consequently, the layer mappings in a DNN with $\mathsf{DropConnect}$ gives rise to the following non-trivial result.

\begin{theorem}[DNN with $\mathsf{DropConnect}$ generalization bound]\label{Thm: dropconnect DNN gen ub}
Consider the DNN model with  $\mathsf{DropConnect}$ probabilities $\delta_{l,i,j}\in(0,1)$, $i=1,\ldots, d_l, j=1,\ldots, d_{l-1}$ at the $l\textsuperscript{th}$ layer (cf.\ \eqref{Eq: dropconnect layer}), $l=1,\ldots,L$. If the loss function $\ell(\bw,X,Y)$ is $\sigma$-sub-Gaussian under $P_{X,Y}$, for all $\bw\in\calW$, we have
\begin{align}
    &| \gen(P_{\bW|D_n},P_{X,Y}) |\\
    &\leq \!\frac{\sigma\sqrt{2}}{n} \! \sum_{i=1}^n\!\sqrt{\prod_{l=1}^{L}\bigg(\! 1-\prod_{i=1}^{d_l}\prod_{j=1}^{d_{l-1}}\delta_{l,i,j}\!\bigg) \sI(X_i;\bW|Y_i) \!+\! \sI(Y_i;\bW) }.
\end{align}
\end{theorem}
The proof of \cref{Thm: dropconnect DNN gen ub} can be derived following the same procedures in that of \cref{Thm: dropout DNN gen ub} and is thus omitted. We observe that the product of coefficients also decrease from $1$ to $0$ as the $\mathsf{DropConnect}$ probabilities increase, the layer dimensions shrink, or the number of layers increases. An exponential decay similar to that discussed after \cref{Thm: dropout DNN gen ub} also occurs in this case. By combining $\mathsf{DropConnect}$ with $\mathsf{Dropout}$ at the input layer $X_i$, a contraction is enforced upon  the mutual information terms $(\sI(X_i;\bW|Y_i), \sI(Y_i;\bW))$, resulting in a generalization bound that decays with the increase of dropout probabilities.

\subsection{Generalization Bound for Noisy DNNs}\label{Sec: gen bound noisy DNN}

Injecting Gaussian noise to neuron outputs is another form of DNN regularization that has been explored in various research studies \cite{neelakantan2015adding,bishop1995training,gitman2019understanding,goldfeld2019estimating}. The idea is to add random Gaussian noise to the input, the network parameters or the internal representation during training. This can help prevent overfitting and improve the model's generalization by introducing a level of uncertainty and robustness \cite{gitman2019understanding}. 
In this section, 
to arrive at nontrivial SDPI contraction coefficient we consider a noisy DNN  model where the feature map at each  layer is perturbed by isotropic Gaussian noise, i.e., 
\begin{equation}
\widetilde{T}_l=  g_{\bW_l}(\widetilde{T}_{l-1})+\epsilon_lZ_l, \quad l=1,\ldots,L,  \label{eq:noisy_DNN}
\end{equation}
where $\widetilde{T}_0=X$, $Z_l\sim N(0,\mathbf{I}_{d_l})$ is independent of the input and $\epsilon_l\in\bbR_{> 0}$ is a constant. The illustration is shown in Figure \ref{Fig: gaussian noise NN}. 
Such noisy DNNs were explored in \cite{goldfeld2019estimating} and were shown to serve as good approximations of classical (deterministic) networks.

To analyze generalization error under the noisy DNN model, we present the following lemma that bounds the SDPI coefficient for the aforementioned channel.

\begin{lemma}[Noisy DNN SDPI coefficient bound]\label{Lem: Y=g(x)+noise coeff}
    Let $X\sim P_X\in\cP(\RR^{d_x})$ and consider a bounded function $g:\RR^{d_x}\to\RR^{d_y}$. Set $Y=g(X)+\epsilon N$, where $\epsilon>0$ and $N\sim\calN(0,\bI_{d_y})$ is independent of $X$. The SDPI coefficient of the induced channel $P_{Y|X}$ satisfies
    \begin{align}
        \eta_f(P_{Y|X})\leq \eta_{\,\TV}(P_{Y|X})\leq 1-2\qfunc\bigg(\frac{\sqrt{2 d_y}\|g\|_\infty}{2 \epsilon}\bigg),
    \end{align}
where $\qfunc(x)\coloneqq\int_x^\infty \frac{1}{\sqrt{2\pi}}e^{-t^2/2} \rmd t$ is the Gaussian complimentary cumulative distribution function.    
\end{lemma}
Lemma \ref{Lem: Y=g(x)+noise coeff}, which is proven in Appendix \ref{App: pf of Lem: Y=g(x)+noise coeff}, implies that whenever $\sqrt{d_y}\|g\|_\infty\epsilon^{-1}>0$, we have $\eta_\KL(P_{Y|X})<1$. Consequently, the layer mappings in a noisy DNN with bounded activations present non-trivial contraction, which gives rise to the following result.

\begin{theorem}[Noisy DNN generalization bound]\label{Thm: noisy DNN gen ub}
Consider the noisy DNN model from \eqref{eq:noisy_DNN} with bounded activation functions $\activ_l, l=1,\ldots,L$. 
If the loss function $\ell(\bw,X,Y)$ is $\sigma$-sub-Gaussian under $P_{X,Y}$, for all $\bw\in\calW$, we have
\begin{align}
    &| \gen(P_{\bW|D_n},P_{X,Y}) |\leq  \!\frac{\sigma\sqrt{2}}{n}\sum_{i=1}^n\\
    &\sqrt{ \prod_{l=1}^{L}\!\!\bigg(\! 1-2\qfunc\bigg(\!\frac{\sqrt{2 d_{l}} \|\phi_l\|_\infty }{2\epsilon_l} \!\bigg)\!\bigg) \sI(X_i;\bW|Y_i)+\sI(Y_i;\bW)}.
\end{align}
\end{theorem}
The proof of Theorem \ref{Thm: noisy DNN gen ub} is provided in Appendix \ref{App: pf of Thm: noisy DNN gen ub}, which follows the similar procedures as in the proof of \cref{Thm: dropout DNN gen ub}. The term $\sI(Y_i;\bW)$ is factored out since the label is not processed by the noisy DNN and upper bounded by $\log K$. The SDPI is invoked $L$ times to arrive at the product of coefficient for the remaining term.  We also note that $\|\activ_l\|_\infty$ is typically small, e.g., $\|\activ_l\|_\infty=1$ if $\activ_l\in\{\mathrm{sigmoid}, \mathrm{softmax} , \tanh\}$. From \cref{Thm: noisy DNN gen ub}, we observe that for fixed noise parameters $\epsilon_1,\ldots,\epsilon_L$, the coefficient product decreases from $1$ to $0$ as the layer dimensions $\{d_l\}_{l=1}^L$ shrink and the number of layers $L$ grows.  If $\min_{l\in[L]}(1-2\qfunc(\frac{\sqrt{2 d_{l}} \|\phi_l\|_\infty }{2\epsilon_l})) \leq q\in (0,1)$ (i.e., $\max_{l\in[L]}d_l\leq (\frac{2\epsilon_l \qfunc^{-1}(\frac{1-q}{2})}{\|\phi_l\|_\infty})^2$), the coefficient is less than $\exp(-L\log \frac{1}{q})$, which decays exponentially in $L$. Furthermore, by increasing the noise level $\epsilon_l$'s, the coefficient product shrinks as well.
The observations are consistent with those of Theorems \ref{Thm: dropout DNN gen ub} and \ref{Thm: dropconnect DNN gen ub}. 

To analyze the behavior of $\sI(X_i;\bW|Y_i)$ and  generalization bounds in the aforementioned  theorems concerning layer dimensions and network depth, we examine the following tractable instance of DNNs. 



\begin{figure}[!t]
    \centering
    \begin{subfigure}[t]{.4\textwidth}
        \centering
        \includegraphics[width=\textwidth]{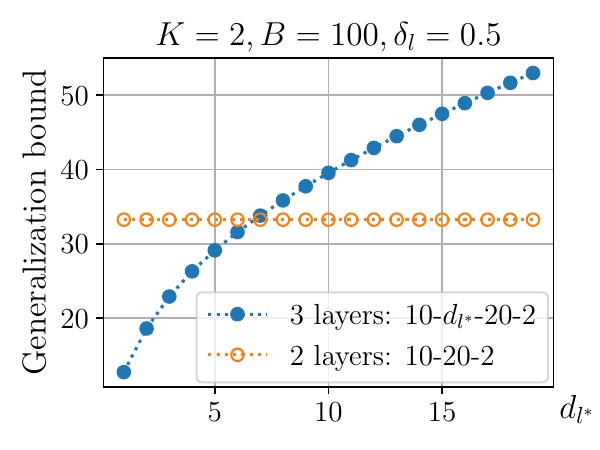}
        \caption{NN with $\mathsf{Dropout}$} 
        \label{fig:finite hypothesis ub dropout}
    \end{subfigure}
    \begin{subfigure}[t]{.4\textwidth}
        \centering
        \includegraphics[width=\textwidth]{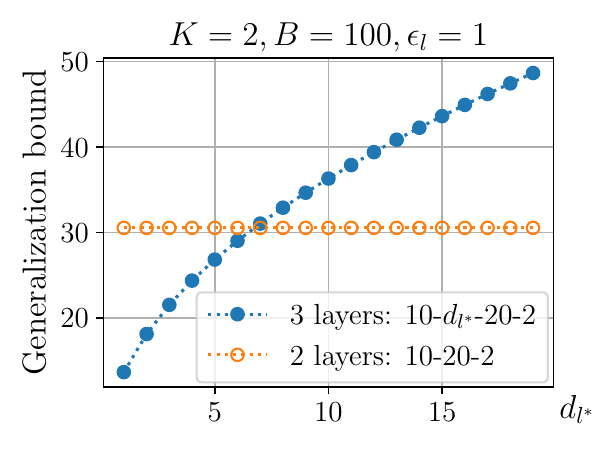}
        \caption{NN with Gaussian noise} 
        \label{fig:finite hypothesis ub}
    \end{subfigure}
    \caption{Examples of tightened generalization bounds for two types of NNs with stochasticity by adding an extra hidden layer.}
\end{figure}
\begin{figure}[!t]
    \centering
    \begin{subfigure}[t]{.4\textwidth}
        \centering
        \includegraphics[width=\textwidth]{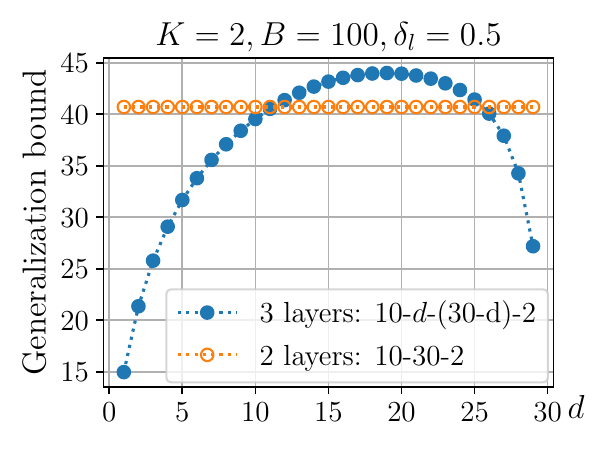}
        \caption{NN with $\mathsf{Dropout}$} 
        \label{fig:finite hypothesis ub dropout division}
    \end{subfigure}
    \begin{subfigure}[t]{.4\textwidth}
        \centering
        \includegraphics[width=\textwidth]{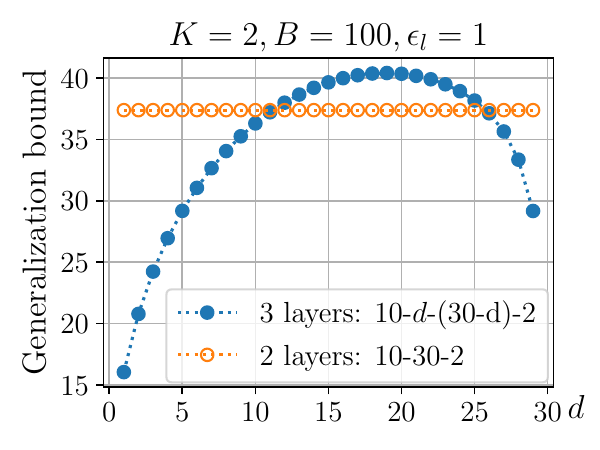}
        \caption{NN with Gaussian noise} 
        \label{fig:finite hypothesis ub division}
    \end{subfigure}
    \caption{Examples of tightened generalization bounds for two types of NNs with stochasticity by dividing one hidden layer into two separate hidden layers.}
\end{figure}
\begin{example}[Numerical evaluation of tightened generalization bounds]
Consider the DNN parameter space is constrained to be finite, e.g., $\calW=[B]^{d_1\times d_0}\times \cdots \times [B]^{d_L\times d_{L-1}}$ for some $B\in\NN$. 
We can upper bound the mutual information by $\sI(X_i;\bW|Y_i)\leq \sH(\bW)\leq\sum_{l=1}^L d_ld_{l-1}\log B$, which increases as $d_l$ and $L$ grow. Note that the equalities hold when the learning algorithm $P_{\bW|D_n}$ is deterministic, and the prior distribution of $\bW$ is uniform. This simplification provides intuition and enables us to visualize the bound.
We consider the following two cases of increasing the depth of an $L$-layer neural net with discrete parameters as above. 

\noindent 1) \underline{Add an extra hidden layer with  dimension $d_{l^*}$.}
We compare a 2-layer neural net with layer dimensions $\{d_0=10,d_1=20,d_2=2\}$ to a 3-layer neural net with layer dimensions $\{d_0=10,d_1=d_{l^*}, d_2=20,d_3=2\}$.
Figures \ref{fig:finite hypothesis ub dropout} and \ref{fig:finite hypothesis ub} plot the said mutual information bounds (ignoring $\sigma\sqrt{2}$ without loss of generality) from Theorems \ref{Thm: dropout DNN gen ub} and  \ref{Thm: noisy DNN gen ub}, respectively. The tightened bounds with the contraction coefficient behave similarly for both the NN with $\mathsf{Dropout}$ and NN with Gaussian noise. We observe that if $d_l^*$ is sufficiently small, the generalization bound shrinks as a result of the added layer.

\medskip

\noindent 2) \underline{Divide the dimension of one hidden layer into two separate} \underline{hidden layers.} We compare a 2-layer neural net with layer dimensions $\{d_0=10,d_1=30,d_2=2\}$ to a 3-layer neural net with layer dimensions $\{d_0=10,d_1=d, d_2=30-d,d_3=2\}$. By varying $d\in[1,29]$, Figures \ref{fig:finite hypothesis ub dropout division} and \ref{fig:finite hypothesis ub division} plot the said mutual information bounds (ignoring $\sigma\sqrt{2}$ without loss of generality) from Theorems \ref{Thm: dropout DNN gen ub} and  \ref{Thm: noisy DNN gen ub}, respectively. The tightened bounds with the contraction coefficient still behave similarly for both the NN with $\mathsf{Dropout}$ and NN with Gaussian noise.
We observe that if one of the splitted layer dimensions, i.e., $d$ or $30-d$ in this example, is sufficiently small, the generalization bound shrinks as a result of the layer division.

\end{example}

  This example suggests that a deep but narrower network may generalize better. 
We observe that when $d_{l^*}$ or $d$ is equal to 1, the generalization bound is minimized. This phenomenon is due to the fact that we are taking a worst-case upper bound on $\sH(\bW)$. Thus, to draw more compelling conclusions, it is necessary to conduct thorough analyses and experiments for general algorithms/architectures.
Next, we prove that the implications from this example can be generalized to the  Gibbs algorithm.

\subsection{Applications to Gibbs Algorithm}
We further characterize the tightened generalization bounds in Theorems \ref{Thm: dropout DNN gen ub}, \ref{Thm: dropconnect DNN gen ub} and \ref{Thm: noisy DNN gen ub}, for DNNs with $\mathsf{Dropout}$, $\mathsf{DropConnect}$ and injected Gaussian noise, under the Gibbs algorithm.  The Gibbs algorithm is a tractable and idealized model for learning algorithms with various regularization approaches, e.g., stochastic optimization methods or relative entropy regularization \cite{raginsky2017non}. This allows us to investigate the concrete impacts of the DNN structure parameters on the mutual information terms in the generalization bounds.

Given any training dataset $D_n=s$, consider the $(\alpha,\pi(\bW),\cL_\sE(\bW,s))$-Gibbs algorithm \cite{aminian2021exact} 
\begin{align}
    P_{\bW|D_n}^\alpha(\bW|s) \coloneqq \frac{\pi(\bW)\exp(-\alpha \cL_\sE(\bW,s))}{\Lambda_{\alpha}(s)}, \label{Eq: gibbs algo}
\end{align}
where $\alpha>0$ denotes the inverse temperature, $\pi\in \calP(\calW)$ is an arbitrary prior distribution on $\bW$ and $\Lambda_{\alpha}(s) = \bbE[\exp(-\alpha \cL_\sE(\bW,s)]$ is the partition function.  The $(\alpha,\pi(\bW),\cL_\sE(\bW,s))$-Gibbs algorithm can be viewed as a randomized version of empirical risk minimization (ERM). As the inverse temperature $\alpha \to \infty$, then network parameters generated by the Gibbs algorithm converges to those from the standard ERM. In \cite{raginsky2017non}, it is proven that under some conditions on the loss function, the posterior $P_{\bW|D_n}$ induced by Stochastic Gradient Langevin Dynamics (SGLD) algorithm \cite{welling2011bayesian} is close to the Gibbs posterior $P_{\bW|D_n}^\alpha$ in 2-Wasserstein distance for sufficiently large iterations.

When DNNs with non-trivial SDPI contraction coefficients (c.f. Lemmas \ref{Lem: dropout SDPI coeff}, \ref{Lem: dropconnect SDPI coeff} and \ref{Lem: Y=g(x)+noise coeff}) are trained with the Gibbs alogrithm using a loss function bounded within $[0,1]$, we obtain the distribution-free bound that decreases as the number of network layers increases.
\begin{proposition}[Tightened generalization bound for Gibbs algorithm]\label{Prop: gibbs ub}
    Consider that the DNN model from \cref{Eq: dropout def} with $\mathsf{Dropout}$ probability $\delta_l\in(0,1)$, \cref{Eq: dropconnect layer} with $\mathsf{DropConnect}$ probability $\delta_{l,i,j}\in(0,1)$, or \cref{eq:noisy_DNN} with bounded activation functions $\activ_l$, $l=1,\ldots,L$, is trained by the $(\alpha,\pi(\bW),\cL_\sE(\bW,D_n))$-Gibbs algorithm with any loss function $\ell: \calW\times \calX \times \calY \to [0,1]$. The SDPI contraction coefficients for these DNN models are denoted by  $\tilde{\eta}_\KL(l)\coloneqq 1-\delta_{l-1}^{d_{l-1}}$, $\tilde{\eta}_\KL(l)\coloneqq 1-\prod_{i=1}^{d_l}\prod_{j=1}^{d_{l-1}}\delta_{l,i,j}$ and $\tilde{\eta}_\KL(l)\coloneqq 1-2\qfunc\big(\frac{\sqrt{2 d_{l}} \|\phi_l\|_\infty }{2\epsilon_l} \big)$, respectively.
    There exists a constant $\gamma\in[0,1]$ such that
    \begin{align}
    \big|\gen(P_{\bW|D_n}^\alpha,P_{X,Y})\big| \leq \frac{\alpha}{4n}\sqrt{ \gamma \prod_{l=1}^{L}\tilde{\eta}_\KL(l)+1-\gamma}.
\end{align}
\end{proposition}
\cref{Prop: gibbs ub}, which is obtained by combining \cite[Theorem 2]{raginsky2016information} and Theorems \ref{Thm: dropout DNN gen ub}, \ref{Thm: dropconnect DNN gen ub} and \ref{Thm: noisy DNN gen ub},  is tighter than \cite[Theorem 2]{raginsky2016information}. Compared to the exact characterization of generalization error in \cite[Theorem 1]{aminian2021exact}, which did not explicitly show the dependence on the neural net structure, \cref{Prop: gibbs ub} shows that the bound is of order $\cO(\frac{1}{n})$ and shrinks with increasing $L$ and decreasing $d_l$ due to the SDPI coefficient $\tilde{\eta}_\KL(l)\in(0,1)$.
As per the analyses presented in Sections \ref{Sec: gen bd for dropout}--\ref{Sec: gen bound noisy DNN}, the generalization bound for the Gibbs algorithm approximately exhibits an exponential decay with increasing network depth $L$ and decreasing layer dimension $d_l$. This observation suggests the benefit of deeper network depth and narrower layer dimensions when the DNN model is trained using regularization techniques, such as $\mathsf{Dropout}$, $\mathsf{DropConnect}$ and noise injection, in conjunction with the Gibbs algorithm. Some other studies have also shown the generalization capability of deep and narrow neural networks \cite{kidger2020universal,kawaguchi2022generalization,zhang2021understanding,he2016deep,lee2022neural}.

\begin{proof}[Proof of \cref{Prop: gibbs ub}]

For any loss function $\ell: \calW\times \calX \times \calY \to [0,1]$, $\ell(\bw,X,Y)$ is $\frac{1}{2}$-sub-Gaussian under $P_{X,Y}$ for any $\bw\in\calW$. Let $D_n^{-i}=D_n\backslash (X_i,Y_i)$ for any $i\in[n]$.
The Gibbs algorithm $P_{\bW|D_n}^\alpha$ with a loss function $\ell\in[0,1]$ is $(\frac{\alpha^2}{8n^2},P_{X,Y})$-stable in mutual information (cf.\ \cite[Definition 1, Theorem 4]{raginsky2016information}), i.e., $\frac{1}{n}\sum_{i=1}^n\sI(\bW;X_i,Y_i|D_n^{-i})\leq \frac{\alpha^2}{8n^2}$. Since $(X_i,Y_i)$ is independent of $D_n^{-i}$, we have $\frac{1}{n}\sum_{i=1}^n\sI(\bW;X_i,Y_i)\leq \frac{1}{n}\sum_{i=1}^n\sI(\bW;X_i,Y_i|D_n^{-i})\leq \frac{\alpha^2}{8n^2}$.
There exists an independent constant $\gamma\in[0,1]$ such that $\frac{1}{n}\sum_{i=1}^n\sI(\bW;X_i|Y_i)\leq \frac{\gamma\alpha^2}{8n^2}$ and $\frac{1}{n}\sum_{i=1}^n\sI(\bW;Y_i)\leq \frac{(1-\gamma)\alpha^2}{8n^2}$.
By further applying the Jensen's inequality to Theorem \ref{Thm: dropout DNN gen ub}, \ref{Thm: dropconnect DNN gen ub} or \ref{Thm: noisy DNN gen ub}, we obtain the bound.
\end{proof}

\section{Concluding Remarks}
This work set to quantify the generalization error of DNNs within the framework of information theory. We derived two hierarchical generalization bounds that capture the effect of depth through the internal representations of the corresponding layers. The two bounds compare the distributions of internal representations of the training and test data under (i) the KL divergence, and (ii) the 1-Wasserstein distance. The KL divergence bound diminishes as the layer index increases, suggesting the advantage of deep network architectures. The Wasserstein bound is minimized by the so-called \emph{generalization funnel layer}, providing new a insight that certain layers play a more prominent role than others in governing generalization performance. We instantiated these results to a binary Gaussian mixture classification task with linear DNNs. Simple analytic expressions for the two generalization bounds we obtained, with the KL divergence reducing to depend on (and shrinks with) the rank of the product of weight matrices, while the Wasserstein bound simplified to depend on the Frobenius norm of the weight matrix product. The latter further implied that the generalization funnel layer of a given model varies with different training methods.

Next, we set to quantify the contraction of the KL divergence bound as a function of depth. To that end, we analyze the SDPI coefficient between  consequent layers in  three regularized DNN models: $\mathsf{Dropout}$, $\mathsf{DropConnect}$, and independent Gaussian noise injection (stochasticity is necessary to avoid a degenerate SDPI coefficient). 
This enables refining our KL divergence generalization bound to capture the contraction through the product of SDPI coefficient associated with different layers, which diminishes from $1$ to $0$ as the depth $L$ increases or the per-layer width $d_l$ shrinks. It was further demonstrated that the rate of decay is approximately exponential and that the SDPI coefficient decays with an increase of network stochasticity, such as higher $\mathsf{Dropout}$/$\mathsf{DropConnect}$ probabilities and higher injected noise level. Numerical evaluation of our bounds are provided for a simple DNN model, where the parameter space is finite. The simulations suggest that, in this example, deeper yet narrower network architectures generalize better. While the extent to which this statement applies generally is unclear, we provide a proof that it is true when learning is performed via the Gibbs algorithm. Overall, our results provide an information-theoretic perspective on generalization of deep models, encompassing quantitative hierarchical bounds to insights into architectures that may generalize better.




Several appealing future research directions extend from this work. 
First, we can generalize our analyses to other neural network architectures such as convolution layer, pooling layer, residual connection, attention layer and so on. It would be interesting to investigate the effects of the properties of different  layer types on  generalization performance. 
Second, tightening SDPI coefficients would be an interesting direction. If we have some prior knowledge about the input distributions, can we improve the SDPI coefficient from the current worst-cast one? It would also be insightful to leverage non-linear SDPI \cite{polyanskiy2017strong} to obtain contraction curve for  deterministic neural networks and a wider range of activation functions.
Third, the behaviour of the mutual information term is still worth investigating. Even though the SDPI coefficient product shrinks with deeper and narrower neural network, the mutual information $\sI(X_i,Y_i;\bW)$ may increase or decrease according to various DNN architectures and training algorithms. We can set out from the stability perspective. By showing the dependence of the stability parameters on DNN architecture, we can make similar analysis to that of the Gibbs algorithm. We can also extend the assumption of  finite parameter space to the setting in the two-stage algorithm \cite[Section 4.2]{xu2017information}, where the parameter space is countable w.r.t.\ the training dataset in the first stage. The mutual information will be upper bounded by a VC dimension, which can be further quantified by  DNN architecture.
Finally, since generalization error tracks the gap between  test and training losses, for non-empricial-risk-minimization algorithms, we need to consider excess risk to better understand the performance of the algorithms. It would be interesting to extend our established tools to bounding  excess risk.
Through these analyses, we will have a clearer view on the impact of DNN architectures on generalization capability.
Moving forward, it would be valuable to conduct thorough analyses and experiments for general algorithms and network structures to draw more compelling conclusions on the impact of network depth and width and find out the generalization funnel layer. 

\appendix

\subsection{Proof of Theorem \ref{Thm: NN gen ub}}\label{App: pf of Thm: NN gen ub}
Let us rewrite the risks and generalization error under the DNN setup. 
Let $(X,Y)\sim  P_{X,Y}$ be a pair of test data sample. At each layer $l$,  the internal representation $T_l$ of a test data feature $X$ is conditionally independent of $W_{l+1}^L$ given $W_{1}^l$.  For any $\bW \in\calW$, let the loss function be rewritten as $\ell(\bW,X,Y)=\ell(g_{\bW_L} \circ g_{\bW_{L-1}}\circ \cdots \circ g_{\bW_1}(X),Y)$.
The expected  population risk over all possible $\bW$ is given by
\begin{align}
    &\bbE_W[\calL_\sP(\bW,P_{X,Y})]\\
    &=\bbE[\ell(g_{\bW_L} \circ g_{\bW_{L-1}}\circ \cdots \circ g_{\bW_1}(X),Y)] \nn\\
    &=\bbE[\ell(g_{\bW_L} \circ g_{\bW_{L-1}}\circ \cdots \circ g_{\bW_{l+1}}(T_l),Y)] \\
    &=\bbE[\bbE[\ell(g_{\bW_L} \circ g_{\bW_{L-1}}\circ \cdots \circ g_{\bW_{l+1}}(T_l),Y)|\bW_1^l]]
\end{align}
where $l\in[L]$ and given $\bW_1^l$, $(T_l,Y)$ are independent of $\bW_{l+1}^L$.

Denote the overall feature mapping function as $f_{\bW}\triangleq g_{\bW_L} \circ g_{\bW_{L-1}}\circ \cdots \circ g_{\bW_1}$. Similarly, for any $l\in[L]$, the expected empirical risk can also be rewritten as
\begin{align}
    &\bbE[\calL_\sE(\bW,D_n)]\\
    &=\bbE\bigg[\frac{1}{n}\sum_{i=1}^n \ell(f_{\bW}(X_i),Y_i) \bigg] =\frac{1}{n}\sum_{i=1}^n\bbE[\ell(f_{\bW}(X_i),Y_i)]\\
    &=\frac{1}{n}\sum_{i=1}^n \bbE[\ell(g_{\bW_L} \circ g_{\bW_{L-1}}\circ \cdots \circ g_{\bW_{l+1}}(T_{l,i}),Y_i)]\\
    &=\frac{1}{n}\sum_{i=1}^n \bbE\big[\bbE[\ell(g_{\bW_L} \circ g_{\bW_{L-1}}\circ \cdots \circ g_{\bW_{l+1}}(T_{l,i}),Y_i)|\bW_1^l]\big].
\end{align}
For notational simplicity, let $g_{\bW_k^j}\coloneqq g_{\bW_k}\circ g_{\bW_{k-1}}\circ \cdots \circ g_{\bW_{j}}$ for any $k<j$ and $k,j\in\bbN$. Then the expected generalization error can be rewritten as
\begin{align}
    &\gen(P_{\bW|D_n},P_{X,Y})
    =\frac{1}{n}\sum_{i=1}^n \bbE\bigg[\bbE[\ell(g_{\bW_{l+1}^L}(T_l),Y)|\bW_1^l]\\
    &\hspace{10em} -\bbE[\ell(g_{\bW_{l+1}^L}(T_{l,i}),Y_i)|\bW_1^l] \bigg]. \label{Eq: NN gen def}
\end{align}

If the loss function $\ell(\bw,X,Y)$ is $\sigma$-sub-Gaussian under $P_{X,Y}$ for all $\bw\in\calW$, we also have for any $l\in[0:L]$, $\ell(g_{\bw_{l+1}^L} (T_l),Y)$ is $\sigma$-sub-Gaussian under $P_{T_l,Y|\bW=\bw}$ for all $\bw\in\calW$.
From Donsker-Varadhan representation, we have for any $\lambda\in\bbR$,
\begin{align}
    &\sD_\KL(P_{\bW_{l+1}^{L},T_{l,i},Y_i|\bW_1^{l}} \| P_{T_l,Y|\bW_1^{l}} \otimes P_{\bW_{l+1}^{L}|\bW_1^{l}}) \nn\\
    &\geq \bbE_{\bW_{l+1}^{L},T_{l,i},Y_i|\bW_1^{l}}[\lambda \ell(g_{\bW_{l+1}^{L}}(T_{l,i}),Y_i)] \\
    &\quad -\log \bbE_{\bW_{l+1}^{L}|\bW_1^{l}}\bbE_{T_l,Y|\bW_1^{l}}[\exp(\lambda \ell(g_{\bW_{l+1}^{L}}(T_l),Y))]\\
    &\geq \lambda (\bbE_{\bW_{l+1}^{L},T_{l,i},Y_i|\bW_1^{l}}[\lambda \ell(g_{\bW_{l+1}^{L}}(T_{l,i}),Y_i)]\\
    &\quad -\bbE_{\bW_{l+1}^{L}|\bW_1^{l}}\bbE_{T_l,Y|\bW_1^{l}}[\ell(g_{\bW_{l+1}^{L}}(T_l),Y)] )-\frac{\lambda^2\sigma^2}{2}.
\end{align}
We can decompose $\sD_\KL(P_{\bW_{l+1}^{L},T_{l,i},Y_i|\bW_1^{l}} \| P_{T_{l},Y|\bW_1^{l}} \otimes P_{\bW_{l+1}^{L}|\bW_1^{l}} | P_{\bW_1^{l}})$ as follows
\begin{align}
    &\sD_\KL(P_{\bW_{l+1}^{L},T_{l,i},Y_i|\bW_1^{l}} \| P_{T_{l},Y|\bW_1^{l}} \otimes P_{\bW_{l+1}^{L}|\bW_1^{l}} | P_{\bW_1^{l}}) \label{Eq: KL BD each layer}\\
    &=\sD_\KL(P_{\bW_{l+1}^{L},T_{l,i},Y_i|\bW_1^{l}}\| P_{T_{l,i},Y_i|\bW_1^{l}} \otimes P_{\bW_{l+1}^{L}|\bW_1^{l}}  | P_{\bW_1^{l}}) \\
    &\quad +\sD_\KL(P_{T_{l,i},Y_i|\bW_1^{l}} \| P_{T_{l},Y |\bW_1^{l}}  | P_{\bW_1^{l}})\\
    &=\sI(T_{l,i},Y_i; \bW_{l+1}^{L}| \bW_1^{l}) + \sD_\KL(P_{T_{l,i},Y_i|\bW_1^{l}} \| P_{T_{l},Y |\bW_1^{l}}  | P_{\bW_1^{l}}).
\end{align}
Thus, we have
\begin{align}
    &\sI(T_{l,i},Y_i; \bW_{l+1}^{L}| \bW_1^{l}) +\sD_\KL(P_{T_{l,i},Y_i|\bW_1^{l}} \| P_{T_{l},Y |\bW_1^{l}}  | P_{\bW_1^{l}}) \nn\\
    &= \sD_\KL(P_{\bW_{l+1}^{L},T_{l,i},Y_i|\bW_1^{l}} \| P_{T_l,Y|\bW_1^{l}} \otimes P_{\bW_{l+1}^{L}|\bW_1^{l}} | P_{\bW_1^{l}}) \\
    &\geq \lambda \Big(\bbE_{\bW_1^{l}}\big[\bbE_{\bW_{l+1}^{L},T_{l,i},Y_i|\bW_1^{l}}[\ell(g_{\bW_{l+1}^{L}}(T_{l,i}),Y_i)] \\
    &\quad -\bbE_{\bW_{l+1}^{L}|\bW_1^{l}}\bbE_{T_l,Y|\bW_1^{l}}[ \ell(g_{\bW_{l+1}^{L}}(T_l),Y)] \big]\Big) - \frac{\lambda^2\sigma^2}{2}.
\end{align}
By  optimizing the RHS over $\lambda>0$ and $\lambda\leq 0$, respectively, we finally obtain
\begin{small}
\begin{align}
    &\Big| \bbE_{\bW_1^{l}}\bbE_{\bW_{l+1}^{L},T_{l,i},Y_i|\bW_1^{l}}[\ell(g_{\bW_{l+1}^{L}}(T_{l,i}),Y_i)]-\\
    &\qquad \bbE_{\bW_1^{l}}\bbE_{\bW_{l+1}^{L}|\bW_1^{l}}\bbE_{T_l,Y|\bW_1^{l}}[\ell(g_{\bW_{l+1}^{L}}(T_l),Y)] \Big| \nn \\
    &\leq \!\! \sqrt{2\sigma^2 \!\big(\sI(T_{l,i},Y_i; \bW_{l+1}^{L}| \bW_1^{l}) \!+\! \sD_\KL(P_{T_{l,i},Y_i|\bW_1^{l}} \| P_{T_{l},Y |\bW_1^{l}}  | P_{\bW_1^{l}}) \big)}, \label{Eq: NN MI ub}
\end{align}
\end{small}
which holds for all $l\in[L]$. Conditioned on $\bW_l$, $T_{l,i}$ and $T_l$ are generated by the same process from $T_{l-1,i}$ and $T_{l-1}$, respectively. By the data-processing inequality, the KL divergence in \eqref{Eq: KL BD each layer} can be bounded as follows:
\begin{align}
    &\sD_\KL(P_{\bW_{l+1}^{L},T_{l,i},Y_i|\bW_1^{l}} \| P_{T_{l},Y|\bW_1^{l}} \otimes P_{\bW_{l+1}^{L}|\bW_1^{l}} | P_{\bW_1^{l}})\\
    &\leq \sD_\KL(P_{\bW_{l+1}^{L},T_{l-1,i},Y_i|\bW_1^{l}} \| P_{T_{l-1},Y|\bW_1^{l}} \otimes P_{\bW_{l+1}^{L}|\bW_1^{l}} | P_{\bW_1^{l}})\\
    &=\sD_\KL(P_{\bW_{l}^{L},T_{l-1,i},Y_i|\bW_1^{l-1}} \| P_{T_{l-1},Y|\bW_1^{l-1}} \otimes P_{\bW_{l}^{L}|\bW_1^{l-1}} | P_{\bW_1^{l-1}})\\
    &\,\, \vdots\\
    &\leq \sD_\KL(P_{X_i,Y_i,\bW_1^L}\| P_{X,Y}\otimes P_{\bW_1^L})\\
    &=\sI(X_i,Y_i;\bW).
\end{align}

Therefore, the expected generalization error in \eqref{Eq: NN gen def} can be upper bounded as follows: 
\begin{align}
   &| \gen(P_{\bW|D_n},P_{X,Y}) | \\
   &\leq \frac{1}{n}\sum_{i=1}^n\sqrt{2\sigma^2 \sD_\KL(P_{T_{L,i},Y_i|\bW} \| P_{T_L ,Y |\bW}  | P_{\bW})}\\
   &\leq \frac{1}{n}\sum_{i=1}^n \Big(2\sigma^2 \big(\sI(T_{L-1,i},Y_i; \bW_L| \bW_1^{L-1}) \\
   &\qquad \qquad + \sD_\KL(P_{T_{L-1,i},Y_i|\bW_1^{L-1}} \| P_{T_{L-1},Y |\bW_1^{L-1}}  | P_{\bW_1^{L-1}}) \big)\Big)^{\frac{1}{2}} \\
   &\,\, \vdots \\
   &\leq \frac{1}{n}\sum_{i=1}^n\Big(2\sigma^2 \big(\sI(T_{l,i},Y_i; W_{l}^{L}| \bW_1^{l}) \\
   &\qquad \qquad + \sD_\KL(P_{T_{l,i},Y_i|\bW_1^{l}} \| P_{T_{l},Y |\bW_1^{l}}  | P_{\bW_1^{l}}) \big)\Big)^{\frac{1}{2}} \label{Eq: NN gen ub l<L-1}\\
   &\,\, \vdots \\
   &\leq \frac{1}{n}\sum_{i=1}^n \Big(2\sigma^2 \big(\sI(T_{1,i},Y_i; W_{2}^{L}| \bW_1) \\
   &\qquad \qquad + \sD_\KL(P_{T_{1,i},Y_i|\bW_1} \| P_{T_{1},Y |\bW_1}  | P_{\bW_1}) \big)\Big)^{\frac{1}{2}}\\
   &\leq \frac{1}{n}\sum_{i=1}^n\sqrt{2\sigma^2 \sI(X_i,Y_i;\bW)}.
\end{align}

\subsection{Proof for \cref{rem:special_cases}}\label{App: pf for discrete latent}
We prove the simplified generalization bound for the discrete latent space case. The information-theoretic quantities in $\UB(l)$ can be upper bounded as follows:
\begin{align}
      &\sI(T_{l,i},Y_i; \bW_{l+1}^{L}| \bW_1^{l})+\sD_\KL(P_{T_{l,i},Y_i|\bW_1^{l}} \| P_{T_{l},Y |\bW_1^{l}}  | P_{\bW_1^{l}})\\
      &=\sI(T_{l,i};\bW_{l+1}^{L}| \bW_1^{l})+\sI(Y_i;\bW_{l+1}^{L}| T_{l,i},\bW_1^{l})\\
      &\quad +\sD_\KL(P_{T_{l,i},Y_i|\bW_1^{l}} \| P_{T_{l},Y |\bW_1^{l}}  | P_{\bW_1^{l}})\\
      &=\sH(T_{l,i}| \bW_1^{l})-\sH(T_{l,i}| \bW_1^{l},\bW_{l+1}^{L})\\
      &\quad +\sH(Y_i| T_{l,i},\bW_1^{l})-\sH(Y_i|T_{l,i},\bW_1^{l},\bW_{l+1}^{L})\\
      &\quad -\sH(T_{l,i},Y_i|\bW_1^l)-\bbE_{P_{T_{l,i},Y_i,\bW_1^{l}}}[\log P_{T_{l},Y |\bW_1^{l}}]\\
    &=\sH(T_{l,i}| \bW_1^{l})-\sH(T_{l,i}| \bW_1^{l},\bW_{l+1}^{L})\\
    &\quad +\sH(Y_i| T_{l,i},\bW_1^{l})-\sH(Y_i|T_{l,i},\bW_1^{l},\bW_{l+1}^{L})\\
      &\quad -\sH(T_{l,i}|\bW_1^l)-\sH(Y_i|T_{l,i},\bW_1^l)\\
      &\quad -\bbE_{P_{T_{l,i},Y_i,\bW_1^{l}}}[\log P_{T_{l},Y |\bW_1^{l}}]\\
      &\overset{\text{(a)}}{\leq} -\bbE_{P_{T_{l,i},Y_i,\bW_1^{l}}}[\log P_{T_{l},Y |\bW_1^{l}}]
  \end{align}
 where (a) follows since $\sH(T_{l,i}| \bW_1^{l},\bW_{l+1}^{L})$, $\sH(Y_i|T_{l,i},\bW_1^{l},\bW_{l+1}^{L})\geq 0$.
 Assuming  $q_l(\bw_1^l)\coloneqq \min_{t\in\calT_l, y\in\calY}P_{T_{l},Y |\bW_1^{l}}(t,y|\bw_1^l)\in \big(0,|\calT_l\times \calY|^{-1}\big)$, 
 then 
    \begin{equation}
        \bbE_{P_{T_{l,i},Y_i,\bW_1^{l}}}[\log P_{T_{l},Y |\bW_1^{l}}]\geq \log \bbE[q_l(\bW_1^l)], \quad \text{and}
    \end{equation}
    \begin{align}
        &\sI(T_{l,i},Y_i; \bW_{l+1}^{L}| \bW_1^{l})+\sD_\KL(P_{T_{l,i},Y_i|\bW_1^{l}} \| P_{T_{l},Y |\bW_1^{l}}  | P_{\bW_1^{l}}) \\
        &\leq \log\frac{1}{\bbE[q_l(\bW_1^l)]}.
    \end{align}

\subsection{Proof of \cref{Thm: wasserstein gen ub}}
\label{App: pf of wasserstein gen ub}
Recall the Kantorovich-Rubinstein duality \cite{villani2021topics}: for any two probability measures $P,Q \in \cP(\cX)$, $ \sW_1(P,Q)=\sup_{f\in \mathrm{Lip}_1(\cX)} \bbE_P[f]-\bbE_Q[f]$, where $\mathrm{Lip}_k(\cX)=\{f \in \{f: \calX\to \bbR\}: |f(x)-f(y)|\leq k\|x-y\|, \forall x,y\in\calX\}$, for any $k\in \bbR_{\geq 0}$

Since $\tloss(g_{\bW_L}\circ \cdots \circ g_{\bW_1}(X),Y)$ is $\rho_0$-Lipschitz in $(g_{\bW_L}\circ \cdots \circ g_{\bW_1}(X),Y)$ and $\activ_l(\cdot)$ is $\rho_l$-Lipschitz,  we have for any $\bw$,
\begin{align}
    &|\ell(\bw,x,y)-\ell(\bw,x',y')|\\
    &=|\tloss(g_{\bw_L}\circ \cdots \circ g_{\bw_1}(x),y)-\tloss(g_{\bw_L}\circ \cdots \circ g_{\bw_1}(x'),y')| \\
    &\leq \rho_0 \|(g_{\bw_L}(g_{\bw_1^{L-1}}(x)),y)-(g_{\bw_L}(g_{\bw_1^{L-1}}(x')),y')\|\\
    &= \rho_0 \sqrt{\big(\activ_L(\bw_L g_{\bw_1^{L-1}}(x))-\activ_L(\bw_Lg_{\bw_1^{L-1}}(x'))\big)^2+(y-y')^2}\\
    &\leq \rho_0 \sqrt{\big(\rho_L \|\bw_L\| \|g_{\bw_1^{L-1}}(x)-g_{\bw_1^{L-1}}(x')\| \big)^2+(y-y')^2 }\\
    &=\rho_0 \Big(\rho_{L}^2 \|\bw_L\|^2\|\activ_{L-1}(\bw_{L-1}g_{\bw_1^{L-2}}(x)) \\
    &\qquad \qquad  -\activ_{L-1}(\bw_{L-1}g_{\bw_1^{L-2}}(x'))\|^2 +(y-y')^2 \Big)^{\frac{1}{2}}\\
    &\leq \rho_0 \Big(\rho_{L}^2\rho_{L-1}^2 \|\bw_L\|^2 \|\bw_{L-1}\|^2 \|g_{\bw_1^{L-2}}(x)-g_{\bw_1^{L-2}}(x')\|^2 \\
    &\qquad \qquad  +(y-y')^2 \Big)^{\frac{1}{2}}\\
    &\vdots \\
    &\leq \rho_0 \sqrt{\prod_{j=1}^{L}\rho_{j}^2\|\bw_j\|^2\|x-x'\|^2 +(y-y')^2 }\\
    &\leq \bar{\rho}_0(\bw) \sqrt{\|x-x'\|^2 +(y-y')^2 } \label{Eq: loss lipshitz bound}
\end{align}
where $\bar{\rho}_0(\bw)\coloneqq \rho_0(1\vee \prod_{j=1}^{L}\rho_{j}\|\bw_j\|)$. 
It means $\ell(\bW,X,Y)$ is $\bar{\rho}_0(\bW)$-Lipschitz in $(X,Y)$ for any $\bW$.
Then we have
\begin{align}
    &\gen(P_{\bW|D_n},P_{X,Y})\\
    &=\frac{1}{n}\sum_{i=1}^n \bbE[\ell(\bW,X,Y)-\ell(\bW,X_i,Y_i)] \\
    &\leq \frac{1}{n}\sum_{i=1}^n \bbE\left[\bar{\rho}_0(\bW)\sW_1(P_{X_i,Y_i|\bW}, P_{X,Y|\bW}) \right]. 
\end{align}
For $l=1,\ldots,L$, similarly, we have $\tloss(g_{\bW_l}\circ \cdots \circ g_{\bW_{l+1}}(T_{l}),Y)$ is $\rho_0(1 \vee \prod_{j=l+1}^L \rho_j\|\bW_j\|)$-Lipschitz in $(T_{l},Y)$ for all $\bW$. Let $\bar{\rho}_l(\bW)=\rho_0(1 \vee \prod_{j=l+1}^L \rho_j\|\bW_j\|)$. Then from the definition in  \eqref{Eq: NN gen def},  we have
\begin{align}
    &\gen(P_{\bW|D_n},P_{X,Y})\\
    &\leq \frac{1}{n}\sum_{i=1}^n \bbE\left[ \bar{\rho}_l(\bW) \sW_1(P_{T_{l,i},Y_i|\bW}, P_{T_l,Y|\bW})\right].
\end{align}
The proof is completed by taking the minimum over $l=0,\ldots,L$.

\vspace{1em}
\noindent\underline{\textbf{Justification of our weaker assumption.}} 
In \cite[Theorem 2]{wang2019information} and \cite[Theorem 1]{rodriguez2021tighter}, they assume that  for any $(x,y)\in\calX \times \calY$, $\ell(\bw,x,y)$ is $\rho$-Lipschitz in $\bw$ for some constant $\rho$. In this paper, we assume that $\tloss:\calY\times\calY\to \bbR_{\geq 0}$, $\tloss(\haty,y)$ is $\rho_0$-Lipschitz in $(\haty,y)$ and $\activ_l:\bbR\to\bbR$, $\activ_l(t)$ is $\rho_l$-Lipschitz in $t$. For example, let the loss function be $l(\bw,x,y)=\tloss(\bw^\intercal x,y)=|\bw^\intercal x-y|$. To satisfy their assumption, it requires that for any $(x,y)$, $\big||\bw^\intercal x-y|-|\bw'^\intercal x-y|\big|\leq |\bw^\intercal x-y-(\bw'^\intercal x-y)|=|(\bw^\intercal -\bw'^\intercal)x|\leq \rho|\bw^\intercal -\bw'^\intercal| $, which does not hold if $\calX$ is unbounded. However, this loss function satisfies our assumption: $\big||\bw^\intercal x-y|-|\bw'^\intercal x'-y'|\big|\leq |\bw^\intercal x-y-(\bw'^\intercal x'-y')|\leq \sqrt{(\bw^\intercal x-\bw'^\intercal x')^2+(y-y')^2}$, which means $\tloss$ is $1$-Lip. Therefore, our assumption accommodates a broader class of loss functions, making it less restrictive.

\subsection{Proof for Remark \ref{Rmk: W comp with KL bd}}
\label{App: pf of Rmk: W comp with KL bd}
Let $\mathsf{diam}(\calX)\coloneqq \sup\{\|x-y\|:x,y\in\calX\}$.
From \cite[Theorem 4]{gibbs2002choosing}, Pinsker's and Bretagnolle-Huber inequalities, for any two probability distributions $\mu,\nu\in\calP(\calX)$,  we have 
\begin{align}
    &\sW_1(\mu,\nu)\leq \mathsf{diam}(\calX)\sD_{\TV}(\mu,\nu)\\
    &\leq \mathsf{diam}(\calX) \sqrt{\bigg( \frac{1}{2}\sD_{\KL}(\mu\|\nu) \wedge   \big(1-\exp(-\sD_{\KL}(\mu\|\nu))\big) \bigg)}.
\end{align}

 From Theorem \ref{Thm: NN gen ub} and \cite{wang2023generalization}, the generalization error can be bounded as follows:
\begin{align}
    &\left|\gen(P_{\bW|D_n},P_{X,Y})\right| \\
    &\leq \frac{A}{n} \sum_{i=1}^n \sD_{\TV}(P_{T_{L,i,Y_i|\bW}},P_{T_{L,Y|\bW}} | P_{\bW}) \leq \UB(L),
\end{align}
where $\UB(L)\coloneqq\frac{\sqrt{2}A}{2n}\sum\limits_{i=1}^n\!\sqrt{\sD_\KL\big(P_{T_{L,i},Y_i|\bW} \big\| P_{T_{L},Y |\bW}  \big| P_{\bW} \big)}$. It can be observed that the total variation distance based bound is tighter under this condition.

Let $l^*\coloneqq \min\limits_{l=0,\ldots, L} \frac{1}{n}\sum_{i=1}^n \bbE[\bar{\rho}_l(\bW) \sW_1(P_{T_{l,i,Y_i|\bW}},P_{T_{l,Y|\bW}} )].$
Then we have
\begin{align}
    &\frac{1 }{n}\sum_{i=1}^n \bbE[\bar{\rho}_{l^*}(\bW) \sW_1(P_{T_{l^*,i,Y_i|\bW}},P_{T_{l^*,Y|\bW}}) ]\\
    &\leq \frac{1 }{n}\sum_{i=1}^n \bbE[\bar{\rho}_{L}(\bW) \sW_1(P_{T_{L,i,Y_i|\bW}},P_{T_{L,Y|\bW}} )]\\
    &= \frac{\rho_0}{n}\sum_{i=1}^n  \sW_1(P_{T_{L,i,Y_i|\bW}},P_{T_{L,Y|\bW}}| P_{\bW} )\\
    &\leq \frac{\rho_0 \mathsf{diam}(\calT_L \times \calY)}{n}\sum_{i=1}^n \sD_{\TV}(P_{T_{L,i,Y_i|\bW}},P_{T_{L,Y|\bW}} | P_{\bW}).
\end{align}
 Under our supervised classification setting,  $\mathsf{diam}(\calT_L \times \calY)=K^2$. Thus, when $\rho_0 K^2 \leq A$,  1-Wasserstein distance based bound is even tighter than the one based on the TV-distance. 

\subsection{Proofs for the case study of binary Gaussian mixture}\label{App: pf of gaussian example KL bound}
To simplify some of the notation ahead, the distribution of a Gaussian random variable $X$ with mean $\mu$ and variance $\sigma^2$ is denoted by $\calN_X(\mu,\sigma^2)$. The subscript stressing the random variable at play is convenient for telling apart the various Gaussian distributions to follow; when there is no ambiguity, the subscript is dropped.

Under the binary Gaussian mixture classification setting,  we first know that the class-conditional feature distribution is $P_{X_i|Y_i=y}=\calN_{X_i}(y\mu_0,\sigma_0^2 \bI_{d_0})$, the class distribution is $P_{Y_i}=\Unif(\{-1,+1\})$, and the prior of $\prodWL^\intercal$ is $P_{\prodWL^\intercal}=\calN_{\prodWL^\intercal}(\mu_0,\frac{\sigma_0^2}{n}\bI_{d_0})$. Given any pair of training data sample $(X_i,Y_i)$, we have
\begin{align}
    &\prodWL^\intercal|(X_i,Y_i) \\
    &= \frac{1}{n}Y_iX_i+\frac{1}{n}\sum_{j\ne i}^n Y_jX_j \sim  \calN_{\prodWL^\intercal}(\mu_{\prodWL|i},\bSigma_{\prodWL|i}),
\end{align}
where $\mu_{\prodWL|i}=\frac{1}{n}Y_iX_i+\frac{n-1}{n}\mu_0$ and $\bSigma_{\prodWL|i}=\frac{n-1}{n^2}\sigma_0^2\bI_{d_0}$.
Then the posterior distribution of $(X_i,Y_i)$ given $\prodWL$ is given by 
\begin{align}
    &P_{X_i,Y_i|\prodWL}\\
    &=\frac{P_{\prodWL^\intercal|X_i,Y_i} \, P_{X_i,Y_i}}{P_{\prodWL}}\\
    &=\frac{\calN_{\prodWL^\intercal}(\mu_{\prodWL|i},\bSigma_{\prodWL|i})}{\calN_{\prodWL^\intercal}(\mu_0,\frac{\sigma_0^2}{n}\bI_{d_0})} \times  \frac{1}{2}\calN_{X_i}(Y_i\mu_0,\sigma_0^2 \bI_{d_0})\\
    &=\frac{1}{2}\calN_{X_i}(Y_i\mu_0,\sigma_0^2 \bI_{d_0})\times C_{i}\calN_{\prodWL^\intercal}\bigg(Y_iX_i,\frac{(n-1)\sigma_0^2}{n}\bI_{d_0} \bigg)\\
    &=\frac{1}{2}\calN_{Y_iX_i}\bigg(\prodWL^\intercal,\frac{(n-1)\sigma_0^2}{n}\bI_{d_0} \bigg),
\end{align}
where $C_i=n^{d_0}\sqrt{(\frac{2\pi \sigma^2}{n^2})^{d_0}}\exp\{\frac{1}{2\sigma_0^2}(Y_iX_i-\mu_0)^\intercal(Y_iX_i-\mu_0)\}$. By integrating $P_{X_i,Y_i|\prodWL}$ over $X_i$, we obtain $P_{Y_i=1|\prodWL}=\frac{1}{2}=P_{Y_i=-1|\prodWL}=\frac{1}{2}$. Thus, for any $(y,\prodwL)$, the learned feature posterior is 
\begin{align}
    P_{X_i|Y_i=y,\prodWL=\prodwL}&=\calN_{yX_i}\bigg(\prodwL^\intercal,\frac{(n-1)\sigma_0^2}{n}\bI_{d_0} \bigg)\\
    &=\calN_{X_i}\bigg(y\prodwL^\intercal,\frac{(n-1)\sigma_0^2}{n}\bI_{d_0} \bigg).
\end{align}
Since $(X_i,Y_i)\to\bW \to \prodWL$ and  $(X_i,Y_i) \to \prodWL \to\bW$ both hold, we can also conclude that   $P_{X_i,Y_i|\bW,\prodWL}=P_{X_i,Y_i|\prodWL}=P_{X_i,Y_i|\bW}$ and $P_{Y_i|\bW,\prodWL}=P_{Y_i|\prodWL}=P_{Y_i|\bW}=\Unif\{-1,+1\}$. 

Next, we compute the divergences between the prior and posterior.

\begin{proof}[Proof of \cref{Prop: Gaussian KL bd}]
For any $l\in[L]$, conditioned on $(Y_i,\bW)$,  the distribution of $T_{l,i}=\prodW X_i$ is Gaussian with the mean and covariance 
\begin{align}
    \bbE[T_{l,i}|Y_i, \bW]&=Y_i\prodW \prodWL^\intercal, \\
    \cov[T_{l,i}|Y_i, \bW]&=\frac{(n-1)\sigma_0^2}{n} \prodW \prodW^\intercal.
\end{align}
Similarly for a test data sample, when conditioned on $(Y,\bW,\prodWL)$, the distribution of $T_l=\prodW X$ is Gaussian with mean and covariance 
\begin{align}
     \bbE[T_l|Y, \bW]&=\bbE[T_l|Y, \bW]=Y_i\prodW \mu_0,\\
     \cov[T_l|Y_l, \bW]&=\sigma_0^2 \prodW \prodW^\intercal.
\end{align}

Note that when $\prodW$ is not a full-rank matrix, the covariance matrices $\cov[T_{l,i}|Y_i, \bW,\prodWL]$ and $\cov[T_l|Y_l, \bW,\prodWL]$ are singular. Thus, the posteriors  $P_{T_{l,i}|Y_i,\bW,\prodWL}$ and $P_{T_l|Y,\bW,\prodWL}$ are the push-forwards of two $d_l$-dimensional nonsingular Gaussian distributions to the lower-dimensional space. In fact, the dimension can be further proven to be $\rank(\prodW)$ via eigendecomposition.

Take the test data sample $(X,Y=1)$ for example in the following. Let $X=\mu_0+ Z$ and the $l\textsuperscript{th}$ representation $T_l=\prodW X=\prodW \mu_0+\prodW Z$, where $Z\sim \calN(0,\sigma_0^2 \bI_{d_0})$ and $\prodW Z\sim \calN(0,\sigma_0^2 \prodW \prodW^\intercal)$ is a singular Gaussian distribution. 
Since $\prodW\prodW^\intercal$ is a $(d_l\times d_l)$ positive semi-definite and symmetric matrix, the eigendecomposition (or singular value decomposition (SVD)) is given by
\begin{align}
    \prodW\prodW^\intercal = \bU \bLambda \bU^\intercal=(\bU \bLambda^{\frac{1}{2}}) (\bU \bLambda^{\frac{1}{2}})^\intercal,
\end{align}
where $\bU$ is a $(d_l\times d_l)$ orthogonal matrix, $\bLambda$ is a $(d_l\times d_l)$ diagonal matrix whose entries are the eigenvalues of $\prodW\prodW^\intercal$. Without loss of generality, we assume that the diagonal values (i.e., eigenvalues) in $\bLambda$ are in descending order. 
Note that the SVD of $\prodW$ is $\prodW=\bU \bS \bV^\intercal$ where $\bS$ is a $(d_l\times d_0)$ diagonal matrix, $\bS \bS^\intercal = \bLambda$, $\bV$ is a $(d_0\times d_0)$ orthogonal matrix.
Construct the following two $d_l$-dimenstional vectors:
\begin{align}
    Z_0&\coloneqq (Z,\underbrace{0,\ldots,0}_{(d_l-d_0) \text{ entries}})^\intercal, \\
    \tilZ &\coloneqq (\underbrace{\tilZ_1,\ldots, \tilZ_{\rank(\prodW)}}_{\sim \calN(0,\sigma_0^2 \bI_{\rank(\prodW)})}, \underbrace{0,\ldots,0}_{(d_l-\rank(\prodW))^\intercal \text{ entries}}).
\end{align}
Since the last $(d_l-\rank(\prodW))$ columns of $\bU \bLambda^{\frac{1}{2}}$ are all zero, the following equalities hold:
\begin{align}
    \prodW Z &\overset{\rmd}{=\joinrel=} \bU \bLambda^{\frac{1}{2}} Z_0 \overset{\rmd}{=\joinrel=} \bU \bLambda^{\frac{1}{2}} \tilZ, \quad  \text{and}\\
    T_l &\overset{\rmd}{=\joinrel=} \prodW \mu_0+\bU \bLambda^{\frac{1}{2}} \tilZ.
\end{align}
Thus, only the a subset of $\rank(\prodW)$ covariates of $T_l$ are effective random variables and  the Gaussian PDF of $T_l$ is on the $\rank(\prodW)$-dimensional space. Let $r_l=\rank(\prodW)$. We can define a restriction of Lebesgue measure to the $\rank(\prodW)$-dimensional affine subspace of $\bbR^{r_l}$ where the Gaussian distribution is supported. With respect to this measure, the distribution of $T_l$ given $(\bW,Y=1)$ has the density of the following motif: for all $t\in\bbR^{d_l}$,
\begin{align}
    &P_{T_l|\bW,Y=1}(t) \\
    &=\frac{\exp\bigg(-\frac{1}{2\sigma_0^2}(t-\prodw\mu_0)^\intercal (\prodW\prodW^\intercal)^\dagger (t-\prodW\mu_0) \bigg)}{\sqrt{(2\pi)^{r_l} \det^*(\sigma_0^2 \prodW\prodW^\intercal )}},    \label{Eq: density of T_l}
\end{align}
where $(\prodW\prodW^\intercal)^\dagger=\bU \bLambda^\dagger \bU^\intercal$ is the generalized inverse of $\prodW\prodW^\intercal$ ,$\bLambda^\dagger$ is the pseudo-inverse of $\bLambda$, and $\det^*$ is the pseudo-determinant. In a similar manner, the density of $T_{l,i}$ can obtained by replacing the mean $\prodW \mu_0$ with $\prodW\prodWL^\intercal$ and $\sigma_0^2$ with $\frac{(n-1)\sigma_0^2}{n}$.


Recall that the KL divergence between any two Gaussian distributions $P=\calN(\mu_p,\bSigma_p), Q=\calN(\mu_q,\bSigma_q)\in \calP(\bbR^k)$ for some $k\in\bbN$ is given by
\begin{align}
    &\sD_\KL(P\|Q)\\
    &=\frac{1}{2} \bigg( \log \frac{\det^*(\bSigma_q)}{\det^*(\bSigma_p)}-k+(\mu_p-\mu_q)^\intercal \bSigma_q^\dagger (\mu_p-\mu_q) \\
    &\qquad \quad +\tr(\bSigma_q^\dagger\bSigma_p)  \bigg).
\end{align}

For $l=1,\ldots,L$, the $\UB(l)$ in Theorem \ref{Thm: NN gen ub} can be written as 
\begin{align}
    &\UB(l) =\frac{\sqrt{2}\sigma}{n}\sum_{i=1}^n\sqrt{ \sD_\KL(P_{T_{l,i},Y_i|\bW} \| P_{T_{l},Y|\bW} | P_{\bW}) }\\
    &=\frac{\sqrt{2}\sigma}{n}\sum_{i=1}^n\sqrt{ \sD_\KL(P_{T_{l,i}|Y_i,\bW}P_{Y_i|\bW} \| P_{T_{l}|Y,\bW}P_{Y} | P_{\bW}) }.
\end{align}

From the probability density function (PDF) of $T_l$ and $T_{l,i}$ (c.f. \eqref{Eq: density of T_l}), the KL divergence term in the upper bound $\UB(l)$ can be rewritten  as: for $l=1,\ldots,L$,
\begin{small}
\begin{align}
    &\sD_\KL(P_{T_{l,i}|Y_i,\bW}P_{Y_i|\bW} \| P_{T_{l}|Y,\bW}P_{Y} | P_{\bW})\\
    &=\frac{1}{2}\sD_\KL(P_{T_{l,i}|Y_i=1,\bW} \| P_{T_{l}| Y=1,\bW}  | P_{\bW}) \\
    &\quad +\frac{1}{2}\sD_\KL(P_{T_{l,i} |Y_i=-1,\bW} \| P_{T_{l} | Y=-1, \bW}  | P_{\bW}) \\
    &=\frac{1}{2}\bbE\Bigg[\bbE \bigg[r_l \bigg(\log \frac{n}{n-1}-1+\frac{n-1}{n}\bigg) \\
    & +\frac{1}{\sigma_0^2}   (\mu_0-\prodWL^\intercal)^\intercal (\prodW)^{\intercal} (\prodW \prodW^\intercal)^\dagger  \prodW (\mu_0-\prodWL^\intercal) \bigg|\bW \bigg] \Bigg] \\
    &=\frac{1}{2}\bbE\Bigg[\bbE \bigg[r_l \bigg(\log \frac{n}{n-1}-\frac{1}{n}\bigg) \\
    & + \frac{1}{\sigma_0^2}   (\mu_0-\prodWL^\intercal)^\intercal \bV \bS^\intercal \bU^\intercal \bU \bLambda^\dagger \bU^\intercal  \bU \bS \bV^\intercal (\mu_0-\prodWL^\intercal) \bigg|\bW \bigg] \Bigg]\\
    &=\frac{1}{2}\bbE\Bigg[\bbE \bigg[r_l \bigg(\log \frac{n}{n-1}-\frac{1}{n}\bigg)+\frac{1}{\sigma_0^2}   \|\mu_0-\prodWL^\intercal\|^2 \bigg|\bW \bigg] \Bigg]\\
    &=\bbE\bigg[\frac{r_l}{2} \bigg(\log \frac{n}{n-1}-\frac{1}{n}\bigg) +\frac{1}{2\sigma_0^2}\bbE\big[\|\mu_0-\prodWL^\intercal\|^2 \big] \bigg] \\
    &=\frac{\bbE[r_l]}{2} \bigg(\log \frac{n}{n-1}-\frac{1}{n}\bigg)+\frac{d_0}{2n},
\end{align}
\end{small}
where the last equality follows
since $\frac{\sqrt{n}}{\sigma_0}(\mu_0-\prodWL^\intercal)\sim \calN(0,\bI_{d_0})$, $\frac{n}{\sigma_0^2}\|\mu_0-\prodWL^\intercal\|^2 \sim \chi_{d_0}^2$ and 
\begin{equation}
    \frac{1}{2\sigma_0^2} \bbE\big[\|\mu_0-\prodWL^\intercal\|^2 \big]=\frac{d_0}{2n}.
\end{equation}

The KL divergence term in the upper bound $\widetilde{\UB}(0)$ can be rewritten  as
\begin{small}
\begin{align}
    &\sD_\KL(P_{X_i, Y_i|\bW} \| P_{X, Y }  | P_{\bW})\\
    &=\bbE\!\Bigg[\!\bbE \bigg[\sD_\KL\bigg(\frac{1}{2}\calN_{X_i}\bigg(-\prodWL^\intercal,\frac{(n-1)\sigma_0^2}{n}\bI_{d_0} \bigg) \bigg\| \frac{1}{2}\calN_{X}(-\mu_0,\sigma_0^2\bI_{d_0} )\!\!\bigg) \\
    &\quad + \sD_\KL\bigg(\frac{1}{2}\calN_{X_i}\bigg(\prodWL^\intercal,\frac{(n-1)\sigma_0^2}{n}\bI_{d_0} \bigg) \bigg\| \frac{1}{2}\calN_{X}(\mu_0,\sigma_0^2\bI_{d_0} ) \bigg)  \bigg|\bW \bigg] \Bigg]\\
    &=\frac{1}{2}\bbE\Bigg[\bbE \bigg[ d_0\bigg(\log \frac{n}{n-1}-1+\frac{n-1}{n}\bigg)+\frac{1}{\sigma_0^2}   \|\mu_0-\prodWL^\intercal\|^2  \bigg|\bW \bigg] \Bigg]\\
    &=\frac{d_0}{2}\bigg(\log \frac{n}{n-1}-\frac{1}{n}\bigg)+\frac{1}{2\sigma_0^2}\bbE\big[\|\mu_0-\prodWL^\intercal\|^2 \big]\\
    &=\frac{d_0}{2}\bigg(\log \frac{n}{n-1}-\frac{1}{n}\bigg)+\frac{d_0}{2n}.
\end{align}
\end{small}

Note that $r_l=\rank(\prodW)\leq \min\{\rank(\bW_l), \rank(\bW_{\otimes (l-1)})\}\leq \rank(\bW_{\otimes (l-1)})=r_{l-1}$, for any $l=2,\ldots,L$, and $r_1\leq \min\{d_0,d_1\}$. 

\end{proof}

\begin{proof}[Proof of \cref{Prop: Gaussian W bd}]
     Since the closed form of $\sW_1$ between two Gaussian distributions is not known but known for $\sW_2$ and $\sW_1(\cdot,\cdot)\leq \sW_2(\cdot,\cdot)$, we consider analysing $\sW_2(P_{T_{l,i,Y_i|\bW}},P_{T_{l,Y|\bW}} | P_{\bW})$ as a surrogate of the upper bound in Theorem \ref{Thm: wasserstein gen ub}.
    Following the proof of \cref{Thm: wasserstein gen ub}, we can obtain
    \begin{align}
        &\gen(P_{\bW|D_n},P_{X,Y})\\
        &=\frac{1}{n}\sum_{i=1}^n \bbE[\ell(\bW,X,Y)-\ell(\bW,X_i,Y_i)] \\
        &=\frac{1}{n}\sum_{i=1}^n \bbE[\bbE[\ell(\bW,X,Y)-\ell(\bW,X_i,Y_i)]|\bW,Y_i] \\
        &\overset{\text(a)}{\leq} \frac{1}{n}\sum_{i=1}^n \bbE\big[\bar{\rho}_0(\bW)\sW_1(P_{X_i|Y_i,\bW}, P_{X|Y,\bW}) \big]\\
        &\leq \frac{1}{n}\sum_{i=1}^n \bbE\big[\bar{\rho}_0(\bW)\sW_2(P_{X_i|Y_i,\bW}, P_{X|Y,\bW}) \big],
    \end{align}
    where (a) follows since $P_{Y_i|\bW}=P_{Y}=\Unif\{-1,+1\}$. Similarly, we also have for all $l=1,\ldots,L$.
    \begin{align}
       \! \gen(P_{\bW|D_n},P_{X,Y}) \!\!\leq\!\!\frac{1}{n}\sum_{i=1}^n \bbE\big[\bar{\rho}_l(\bW)\sW_2(P_{T_{l,i}|Y_i,\bW}, P_{T_l|Y,\bW}) \big].
    \end{align}
    By plugging the  $P_{T_{l,i}|Y_i,\bW}$ and $P_{T_l|Y,\bW}$ (c.f. \eqref{Eq: density of T_l}) into the upper bound, we have
    \begin{small}
    \begin{align}
        &\sW_2(P_{T_{l,i}|Y_i, \bW},P_{T_{l}|Y, \bW} )\\
        &=\bigg(\|\prodW (\prodWL^\intercal -\mu_0)\|^2\\
        &\qquad +\tr\Big(\frac{(n-1)\sigma_0^2}{n}  (\prodW \prodW^\intercal)+\sigma_0^2 (\prodW \prodW^\intercal) \\
        &\qquad -2\Big(\frac{(n-1)\sigma_0^4}{n} (\prodW \prodW^\intercal)^{\frac{1}{2}}(\prodW \prodW^\intercal)(\prodW \prodW^\intercal)^{\frac{1}{2}}  \Big)^\frac{1}{2} \Big) \!\! \bigg)^{\frac{1}{2}} \\
        &=\sqrt{\|\prodW (\prodWL^\intercal -\mu_0)\|^2+\Big(\frac{\sqrt{n-1}}{\sqrt{n}}-1 \Big)^2 \sigma_0^2\tr\big( \prodW \prodW^\intercal \big) } \\
        &=\sqrt{\|\prodW (\prodWL^\intercal -\mu_0)\|^2+\frac{(\sqrt{n-1}-\sqrt{n})^2\sigma_0^2 }{n}\|\prodW\|_\rmF^2 }. 
    \end{align}
    \end{small}
    Similarly, we have
    \begin{align}
        &\sW_2(P_{X_{i}|Y_i, \bW},P_{X|Y, \bW} )\\
        &=\sqrt{\|(\prodWL^\intercal -\mu_0)\|^2+ \frac{d_0(\sqrt{n-1}-\sqrt{n})^2\sigma_0^2 }{n}  } .
    \end{align}

    Since the activation functions are all $1$-Lipschitz, i.e., $\rho_l=1$ for all $l=1,\ldots,L$ and the loss function $\tloss$ is $4\sqrt{2}$-Lipschitz, we have $\bar{\rho}_l(\bW)=4\sqrt{2}\big(1\vee \prod_{j=l+1}^L \|\bW_j\|_{\mathrm{op}} \big)$ for $l=0,1,\ldots,L$. 
    
    For notational simplicity, let $\bW_{\otimes 0}=\bW_0=\bI_{d_0}$ and $r_0=d_0$. Here we use $\|\cdot\|_\rmF$ of a vector to equivalently denote its Euclidean norm, with a slight abuse of notations.
    Then the generalization error is upper bounded by
    \begin{strip}
    \par\noindent\rule{\dimexpr(0.5\textwidth-0.5\columnsep-0.4pt)}{0.4pt}%
    \rule{0.4pt}{6pt}
    \begin{align}
        &\gen(P_{\bW|D_n},P_{X,Y})\\
        &\leq \min\bigg\{\min_{l=1,\ldots, L} \bbE\bigg[\bar{\rho}_l(\bW)\sqrt{\|\prodW (\prodWL^\intercal -\mu_0)\|^2+\frac{(\sqrt{n-1}-\sqrt{n})^2\sigma_0^2 }{n}\|\prodW\|_\rmF^2 } \bigg]\; , \\
        &\qquad \qquad \qquad  \bbE\bigg[\bar{\rho}_0(\bW)  \sqrt{\|(\prodWL^\intercal -\mu_0)\|^2+ \frac{d_0(\sqrt{n-1}-\sqrt{n})^2\sigma_0^2 }{n}  } \, \bigg]   \bigg\}\\
        &\overset{\text{(a)}}{\leq} \min\bigg\{\min_{l=1,\ldots, L} \bbE\bigg[\bar{\rho}_l(\bW)\bigg(\|\prodW (\prodWL^\intercal -\mu_0)\|+\frac{(\sqrt{n}-\sqrt{n-1})\sigma_0 }{\sqrt{n}}\|\prodW\|_\rmF \bigg) \bigg]\; , \\
        &\qquad \qquad \qquad  \bbE\bigg[\bar{\rho}_0(\bW)  \bigg( \|(\prodWL^\intercal -\mu_0)\|+ \frac{\sqrt{d_0}(\sqrt{n}-\sqrt{n-1})\sigma_0 }{\sqrt{n}}  \bigg) \, \bigg]   \bigg\}\\
        &\overset{\text{(b)}}{\leq} \min\bigg\{\min_{l=1,\ldots, L} \bbE\bigg[\bar{\rho}_l(\bW)\|\prodW \|_\rmF \bigg(\|(\prodWL^\intercal -\mu_0)\|+\frac{(\sqrt{n}-\sqrt{n-1})\sigma_0 }{\sqrt{n}} \bigg) \bigg]\; , \\
        &\qquad \qquad \qquad  \bbE\bigg[\bar{\rho}_0(\bW)\sqrt{d_0}  \bigg( \|(\prodWL^\intercal -\mu_0)\|+ \frac{(\sqrt{n}-\sqrt{n-1})\sigma_0 }{\sqrt{n}}  \bigg) \, \bigg]   \bigg\}\\
        &= \min_{l=0,\ldots, L} \bbE\bigg[\bar{\rho}_l(\bW)\|\prodW \|_\rmF \bigg(\|\prodWL^\intercal -\mu_0\|+\frac{(\sqrt{n}-\sqrt{n-1})\sigma_0 }{\sqrt{n}} \bigg) \bigg] \\
        &\overset{\text{(c)}}{\leq} \min_{l=0,\ldots, L} \bbE\left[ \bar{\rho}_l(\bW)^2\|\prodW \|_\rmF^2 \right]^{\frac{1}{2}} \bbE\bigg[\bigg(\|\prodWL^\intercal -\mu_0\|+\frac{(\sqrt{n}-\sqrt{n-1})\sigma_0 }{\sqrt{n}} \bigg)^2 \bigg]^{\frac{1}{2}}\\
        &\overset{\text{(d)}}{\leq}\min_{l=0,\ldots, L} \bbE\left[ \bar{\rho}_l(\bW)^2\|\prodW \|_\rmF^2 \right]^{\frac{1}{2}} \bigg(\frac{\sqrt{d_0}\sigma_0}{\sqrt{n}} +\frac{(\sqrt{n}-\sqrt{n-1})\sigma_0 }{\sqrt{n}} \bigg) \\
        &=\frac{4\sqrt{2}\sigma_0(\sqrt{d_0}+(\sqrt{n}-\sqrt{n-1})) }{\sqrt{n}} \min_{l=0,\ldots, L} \bbE\left[ \bigg(1\vee \prod_{j=l+1}^L \|\bW_j\| \bigg)^2\|\prodW \|_\rmF^2 \right]^{\frac{1}{2}},
    \end{align}
    \vspace{\belowdisplayskip}\hfill\rule[-6pt]{0.4pt}{6.4pt}%
    \rule{\dimexpr(0.5\textwidth-0.5\columnsep-1pt)}{0.4pt}
    \end{strip}
    where (a) follows since $\sqrt{a^2+b^2}\leq |a|+|b|$, (b) follows from the Cauchy-Schwarz inequality, (c) follows from the H\"{o}lder's inequality, and (d) follows from Minkowski's inequality and  $\frac{n}{\sigma_0^2}\|\prodWL^\intercal-\mu_0\|^2 \sim \chi_{d_0}^2$. 
\end{proof}

\begin{proof}[Proof of \cref{Ex: eg for Gaussian W bd}]
 Since $\bW_l$ is $(2\times 2)$ rotation matrix multiplied by a scalar factor $C_l$ and $W_L=(0,C_L)$ is a row vector, $\|\bW_l\|_{\mathrm{op}}=C_l$ for $l=1,\ldots,L$. We have $\bar{\rho}_l(\bW)=4\sqrt{2}\big(1\vee \prod_{j=l+1}^L \|\bW_j\|_{\mathrm{op}} \big)=4\sqrt{2}\big(1\vee \prod_{j=l+1}^L C_j \big)$ for $l=0,1,\ldots,L$. 
\end{proof}

\subsection{Proof of \cref{Lem: dropout SDPI coeff}}\label{App: pf of dropout}

Recall the definition of $\mathsf{Dropout}$ technique in \eqref{Eq: dropout def}. Fix any $l\in\{0,\ldots, L\}$.
Given $\bW$, the Markov chain $T_l\to \widetilde{T}_l\to T_{l+1}$ holds and we have the following SDPI:
\begin{align}
    &\sD_\KL(P_{T_{l+1,i}|Y_i,\bW}\| P_{T_{l+1}|Y,\bW} | P_{Y_i,\bW})\\
    &\leq \bbE\big[\bbE[\eta_\KL(P_{T_{l+1}|\widetilde{T}_l,\bW}) \sD_\KL(P_{\widetilde{T}_{l,i}|Y_i,\bW}\| P_{\widetilde{T}_{l}|Y,\bW} ) | Y_i,\bW] \big]\\
    &\leq \eta_\KL(P_{\widetilde{T}_l|T_l}) \bbE\big[\bbE[\eta_\KL(P_{T_{l+1}|\widetilde{T}_l,\bW}) \\
    &\quad \times \sD_\KL(P_{T_{l,i}|Y_i,\bW}\| P_{T_{l}|Y,\bW} ) | Y_i,\bW] \big]\\
    &\overset{\text{(a)}}{=}  \eta_\KL(P_{\widetilde{T}_l|T_l}) \sD_\KL(P_{T_{l,i}|Y_i,\bW}\| P_{T_{l}|Y,\bW} | P_{Y_i,\bW})
\end{align}
where (a) follows from $\eta_\KL(P_{T_{l+1}|\widetilde{T}_l,\bW})=1$ since $P_{T_{l+1}|\widetilde{T}_l,\bW}$ is a deterministic mapping.
We observe that $P_{\widetilde{T}_l|T_l}$ is composed of $d_l$ parallel identical channels and thus, let $P_{\widetilde{T}_l|T_l} \coloneqq P_{\widetilde{T}|T}^{\otimes d_l}$, where $P_{\widetilde{T}|T}$ denotes the channel for $\widetilde{T}=T\cdot E $ with $E\sim \mathsf{Bern}(1-\delta_l)$. Let $\widetilde{\calT},\calT \subset \bbR$ be the alphabets for $\widetilde{T}, T$, respectively. In the following, to avoid any confusion, let $P_{\widetilde{T}|T}^{\delta_l}$ denote $P_{\widetilde{T}|T}$ with cross probability $1-\delta_l$.

\paragraph{Uppder bound for $\eta_\KL(P_{\widetilde{T}_l|T_l})$}
 By applying \cite[Theorem 5]{polyanskiy2017strong} $d_l$ times, we have
\begin{align}
    \eta_\KL(P_{\widetilde{T}_l|T_l})\leq 1-(1-\eta_\KL(P_{\widetilde{T}|T}^{\delta_l}))^{d_l}. \label{Eq: eta_KL parallel channels}
\end{align}
If $T\ne 0$, the channel $P_{\widetilde{T}|T}^{\delta_l}$ is equivalent to the erasure channel with an erasure probability $\delta_l$. 
From \cite{ordentlich2021strong}, which shows that the SDPI  coefficient is achieved by binary inputs, we have
\begin{align}
    \eta_\KL(P_{\widetilde{T}|T}^{\delta_l})&=\sup_{t,t'\in\calT} \sup_{P,Q \in\calP(\{t,t'\})}\frac{\sD_\KL(P_{\widetilde{T}|T}^{\delta_l}\circ P \| P_{\widetilde{T}|T}^{\delta_l}\circ Q)}{\sD_\KL(P\|Q)} \\
    &= \eta_\KL(\mathsf{BEC}(\delta_l)),
\end{align}
If $t t'\ne 0$, the channel $P_{\widetilde{T}|T}^{\delta_l}:\{t,t'\}\to \{t,t',0\}$ is equivalent to the binary the erasure channel with an erasure probability $\delta_l$, denoted as $\mathsf{BEC}(\delta_l)$. If $t t'= 0$, the channel $P_{\widetilde{T}|T}^{\delta_l}:\{t,t'\}\to \{t,t'\}$ is equivalent to the $\mathsf{Z}$-channel with a cross probability $\delta_l$, denoted as $\mathsf{ZC}(\delta_l)$. Thus, the SDPI coefficient is equal to
\begin{align}
    \eta_\KL(P_{\widetilde{T}|T}^{\delta_l})=\eta_\KL(\mathsf{BEC}(\delta_l)) \vee\eta_\KL(\mathsf{ZC}(\delta_l)) . \label{Eq: eta_KL is maximum of BEC and ZC}
\end{align}
 Given the $\mathsf{BEC}(\delta_l)$, for any two binary input distributions $\mathsf{Bern}(\alpha)$ and $\mathsf{Bern}(\beta)$, we have
\begin{align}
    &D_\KL(\mathsf{BEC}(\delta_l) \circ \mathsf{Bern}(\alpha) \| \mathsf{BEC}(\delta_l) \circ \mathsf{Bern}(\beta) ) \\
    &=(1-\delta_l)D_\KL(\mathsf{Bern}(\alpha)\|\mathsf{Bern}(\beta)), \quad \forall \alpha,\beta\in[0,1].
\end{align}
 Thus, $\eta_\KL(\mathsf{BEC}(\delta_l))=1-\delta_l$.

The SDPI coefficient for $\mathsf{ZC}(\delta_l)$ is upper bounded by the Dobrushin's coefficient:
\begin{align}
    \eta_\KL(\mathsf{ZC}(\delta_l)) &\leq \eta_\TV(\mathsf{ZC}(\delta_l)) \\
    &=  \frac{1}{2}(|1-\delta_l|+|0-(1-\delta_l)|)=1-\delta_l.
\end{align}

Therefore, from \eqref{Eq: eta_KL is maximum of BEC and ZC}, $\eta_\KL(P_{\widetilde{T}|T}^{\delta_l})=1-\delta_l$. By plugging $\eta_\KL(P_{\widetilde{T}|T})$ back to \eqref{Eq: eta_KL parallel channels}, we get
 \begin{equation}
    \eta_\KL(P_{\widetilde{T}|T}^{\delta_l})\leq 1-\delta_l^{d_l}. \label{Eq:eta BEC^d ub}
 \end{equation}

\paragraph{Lower bound for $\eta_\KL(P_{\widetilde{T}_l|T_l})$}
Let $\calT_l \subset \bbR^{d_l}$ be the alphabet of $T_l$ and $\calT_l^{-0} \subset (\bbR \backslash \{0\})^{d_l}$. Let $P_{\widetilde{T}_l|T_l}^{\delta_l}$ denote $P_{\widetilde{T}_l|T_l}$ with dropout rate $\delta_l$.
From \cite{ordentlich2021strong}, we have
\begin{align}
    \eta_\KL(P_{\widetilde{T}_l|T_l}^{\delta_l})&\geq \frac{1}{2}\sup_{t,t'\in\calT_l}\sD_{\sH^2}(P_{\widetilde{T}_l|T_l=t}^{\delta_l},P_{\widetilde{T}_l|T_l=t'}^{\delta_l}) \\
    &\geq \frac{1}{2}\sup_{t,t'\in\calT_l^{-0}}\sD_{\sH^2}(P_{\widetilde{T}_l|T_l=t}^{\delta_l},P_{\widetilde{T}_l|T_l=t'}^{\delta_l})
\end{align}
where $\sD_{\sH^2}$ denotes the Hellinger distance.  Given any $t\in\calT_l^{-0}$, the output $\widetilde{T}_l$ follows the binomial distribution $\mathsf{Binom}(d_l,\delta_l)$ on the alphabet $\widetilde{\calT}_l(t)$, where $\{\mathbf{0}\}\subset \widetilde{\calT}_l(t)$. It is straightforward that $\sup_{t,t'\in\calT_l^{-0}}\sD_{\sH^2}(P_{\widetilde{T}_l|T_l=t},P_{\widetilde{T}_l|T_l=t'}) = \\ \sup_{t,t'\in\calT_l^{-0}: \mathsf{Hamming}(t,t')=d_l}\sD_{\sH^2}(P_{\widetilde{T}_l|T_l=t},P_{\widetilde{T}_l|T_l=t'})$. Note that for any $t,t'\in\calT_l^{-0}$ such that $\mathsf{Hamming}(t,t')=d_l$, we have $(\widetilde{\calT}_l(t)\backslash \{\mathbf{0}\})\cap (\widetilde{\calT}_l(t')\backslash \{\mathbf{0}\})=\emptyset$.
Then the corresponding Hellinger distance is given by
\begin{align}
    &\sD_{\sH^2}(P_{\widetilde{T}_l|T_l=t}^{\delta_l},P_{\widetilde{T}_l|T_l=t'}^{\delta_l})\\
    &=\sum_{u \in \widetilde{\calT}_l(t)\cup \widetilde{\calT}_l(t') } \Big(\sqrt{P_{\widetilde{T}_l|T_l=t}^{\delta_l}(u)}-\sqrt{P_{\widetilde{T}_l|T_l=t'}^{\delta_l}(u)} \Big)^2\\
    &=\Big(\sqrt{P_{\widetilde{T}_l|T_l=t}^{\delta_l}(\mathbf{0})}-\sqrt{P_{\widetilde{T}_l|T_l=t'}^{\delta_l}(\mathbf{0})} \Big)^2 \\
    &\qquad + \sum_{u \in \widetilde{\calT}_l(t)\backslash\{\mathbf{0}\}} P_{\widetilde{T}_l|T_l=t}^{\delta_l}(u)+\sum_{u \in \widetilde{\calT}_l(t')\backslash\{\mathbf{0}\}} P_{\widetilde{T}_l|T_l=t'}^{\delta_l}(u)\\
    &\overset{\text{(a)}}{=} 2\sum_{k=0}^{d_l-1} \binom{d_l}{k} \delta_l^k(1-\delta_l)^{n-k}\\
    &=2(1-\delta_l^{d_l}),
\end{align}
where (a) follows since $P_{\widetilde{T}_l|T_l=t}^{\delta_l}(\mathbf{0})=P_{\widetilde{T}_l|T_l=t'}^{\delta_l}(\mathbf{0})=\delta_l^{d_l}$. The lower bound for $\eta_\KL(P_{\widetilde{T}_l|T_l}^{\delta_l})$ is then given by
\begin{align}
    \eta_\KL(P_{\widetilde{T}_l|T_l}^{\delta_l})\geq 1-\delta_l^{d_l}. \label{Eq:eta BEC^d lb}
\end{align}
By combining \eqref{Eq:eta BEC^d ub} and \eqref{Eq:eta BEC^d lb}, we conclude that $\eta_\KL(P_{\widetilde{T}_l|T_l})= 1-\delta_l^{d_l}$.

\subsection{Proof of \cref{Thm: dropout DNN gen ub}}\label{App: pf dropout gen ub}
From the proof of \cref{Lem: dropout SDPI coeff} in \cref{App: pf of dropout}, we have
\begin{align}
    &\sD_\KL(P_{T_{l+1,i}|Y_i,\bW}\| P_{T_{l+1}|Y,\bW} | P_{Y_i,\bW}) \\
    &\leq (1-\delta_l^{d_l}) \sD_\KL(P_{T_{l,i}|Y_i,\bW}\| P_{T_{l}|Y,\bW} | P_{Y_i,\bW}), \; l=0,\ldots,L. \label{Eq: BEC^d SDPI}
\end{align}

   Recall the upper bound \eqref{Eq: gen ub last layer} in Theorem \ref{Thm: NN gen ub}:
\begin{align}
    &| \gen(P_{\bW|D_n},P_{X,Y}) | \\
    &\leq \frac{1}{n}\sum_{i=1}^n\sqrt{2\sigma^2 \sD_\KL(P_{T_{L,i},Y_i|\bW} \| P_{T_{L} ,Y |\bW}  | P_{\bW})}\\
    &=\frac{1}{n}\sum_{i=1}^n\sqrt{2\sigma^2 (\sD_\KL(P_{T_{L,i}|Y_i,\bW} \| P_{T_{L}  |Y,\bW}  | P_{Y_i,\bW})+\sI(Y_i;\bW))}, \label{Eq: noisy NN gen ub middle step}
\end{align}
where \eqref{Eq: noisy NN gen ub middle step} follows since for any $l\in[L]$,
\begin{align}
    &\sD_\KL(P_{T_{l,i},Y_i|\bW} \| P_{T_{l},Y|\bW} | P_{\bW})\\
    &=\sD_\KL(P_{T_{l,i}|Y_i,\bW} P_{Y_i|\bW} \| P_{T_{l}|Y,\bW}P_{Y|W} | P_{\bW})\\
    &=\sD_\KL(P_{T_{l,i}|Y_i,W} \| P_{T_{l}|Y,\bW} | P_{Y_i,\bW})+\sD_\KL(P_{Y_i|\bW} \|P_{Y} | P_{\bW} )\\
    &=\sD_\KL(P_{T_{l,i}|Y_i,W} \| P_{T_{l}|Y,\bW} | P_{Y_i,\bW})+\sI(Y_i;\bW).
\end{align}
It can be observed that the data-processing inequality is only applied to $\sD_\KL(P_{T_{l,i}|Y_i,W} \| P_{T_{l}|Y,\bW} | P_{Y_i,\bW})$ since $Y_i$ is not processed. Furthermore, since $Y_i\in\calY=[K]$ is a discrete random variable, we have $\sI(Y_i;\bW)\leq \sH(Y_i) \leq \log K$.

By induction, if we have $L$ layers
\begin{align}
    &\sD_\KL(P_{T_{L,i}|Y_i,\bW}\| P_{T_{L}|Y,\bW} | P_{Y_i,\bW})\\
    &\leq (1-\delta_{L-1}^{d_{L-1}})\sD_\KL(P_{T_{L-1,i}|Y_i,\bW}\| P_{T_{L-1}|Y,\bW} | P_{Y_i,\bW})\\
    &\leq \prod_{l=L-2}^{L-1}(1-\delta_l^{d_l})  \sD_\KL(P_{T_{L-2,i}|Y_i,\bW}\| P_{T_{L-2}|Y,\bW} | P_{Y_i,\bW})\\
    &\vdots\\
    &\leq \prod_{l=0}^{L-1}(1-\delta_l^{d_l})  \sD_\KL(P_{T_{0,i}|Y_i,\bW}\| P_{T_{0}|Y,\bW} | P_{Y_i,\bW})\\
    &=\prod_{l=1}^L(1-\delta_l^{d_l}) \sI(X_i;\bW|Y_i). \label{Eq: gen ub induction dropout}
\end{align}
The proof is completed by plugging \eqref{Eq: gen ub induction dropout}  into \eqref{Eq: noisy NN gen ub middle step}.


\subsection{Proof of \cref{Coro: Dropout bd mono-shrink}}\label{App: pf of Coro: Dropout bd mono-shrink}
In this proof, we show that the mutual information terms decay as the $\mathsf{Dropout}$ probability $\delta_0$ increases.  Let $\calJ(k)\subseteq [d]$ be a set of coordinate indices, $|\calJ(k)|=k\in[d]$, and $X_i(\calJ(k))=\{X_i(j)\}_{j\in\calJ(k)}$ be a subset of coordinates of $X_i$. 
Some coordinates of the input $X_i$ are randomly deactivated as $0$ before being passed into the neural network.
Consider an independent binary mask vector $E_t\sim \mathsf{Bern}(1-\delta_0)^{\otimes d_0}$ for the input $X_i$. 
It can be observed that $X_i\odot E_t$ is a sufficient statistic of $X_i$ for $\bW$. We decompose the mutual information term as
\begin{align}
    \sI(X_i,Y_i;\bW)
    &=\sI(X_i \odot E_t,Y_i;\bW)\\
    &\overset{(a)}{=}\sum_{e\in \{0,1\}^{d_0}} P_{E_t}(e) \sI(X_i \odot e,Y_i;\bW)
\end{align}
where (a) follows since $E_t$ is independent of $(X_i,Y_i)$. Let $e_{\calJ(k)}\in \{0,1\}^{d_0}$ denote the binary vector with non-zeros elements at indices $\calJ(k)$, for  $k=0,\ldots, d_0$. 
Then $P_{E_t}(e_{\calJ(k)})=\delta_0^{d_0-k}(1-\delta_0)^{k}$. 

Let $I_k^{(i)}=\max_{\calJ(k)\subseteq [d], |\calJ(k)|=k} \sI(X_i(\calJ(k)),Y_i;\bW)$, for $k=0,\ldots,d_0$. We have $0=I_0^{(i)}\leq I_1^{(i)} \leq \ldots \leq I_{d_0}^{(i)}=\sI(X_i,Y_i;\bW)$. Thus, we have
\begin{align}
    &\sum_{e\in \{0,1\}^{d_0}} P_{E_t}(e) \sI(X_i \odot e,Y_i;\bW)\\
    &= \sum_{k=0}^{d_0}\sum_{\calJ(k) \subseteq [d]} \delta_0^{d_0-k}(1-\delta_0)^{k}\sI(X_i(\calJ(k)),Y_i;\bW)\\
    &= \sum_{k=0}^{d_0} \delta_0^{d_0-k}(1-\delta_0)^{k} \sum_{\calJ(k) \subseteq [d]} \sI(X_i(\calJ(k)),Y_i;\bW)  \\
    & \leq \sum_{k=0}^{d_0} \delta_0^{d_0-k}(1-\delta_0)^{k}\binom{d_0}{k} I_k^{(i)}.  \label{Eq: dropout MI ub}
\end{align}
        Let $X,Y,Z$ be random variables following the laws:
        \begin{align}
            \Pr(X=k)&=\Pr(Y=I^{(i)}_k)\\
            &=\Pr(Z=kI^{(i)}_k)\\
            &=\binom{n}{k}(1-\delta_0)^k\delta_0^{d_0-k}, \forall k=0,1,\ldots,d_0,
        \end{align}
        and their expectations are
        \begin{align}
            \bbE[X]&=\sum_{k=0}^{d_0} \binom{n}{k}(1-\delta_0)^k\delta_0^{d_0-k}k,\\
            \bbE[Y]&=\sum_{k=0}^{d_0} \binom{n}{k}(1-\delta_0)^k\delta_0^{d_0-k}I^{(i)}_k, \\
            \bbE[Z]&=\sum_{k=0}^{d_0} \binom{n}{k}(1-\delta_0)^k\delta_0^{d_0-k}kI^{(i)}_k.
        \end{align}
        We notice that $P_{Y|X=k}=\mathbbm{1}\{Y=I^{(i)}_k\}$ and so $\bbE[Z]=\bbE[XY]$. We want to prove that $\bbE[Y]$ decreases as $\delta_0$ increases. Let us take the derivative of $\bbE[Y]$ w.r.t.\ $\delta_0$:
 \begin{align}
            \frac{\partial \bbE[Y]}{\partial \delta_0}&=-\frac{\partial \bbE[Y]}{\partial (1-\delta_0)}\\
            &=-\sum_{k=0}^{d_0}\frac{k}{\delta_0(1-\delta_0)}\binom{d_0}{k}(1-\delta_0)^k\delta_0^{d_0-k}I_k^{(i)} \\
            &\qquad +\sum_{k=0}^{d_0}\frac{n}{\delta_0}\binom{d_0}{k}(1-\delta_0)^k\delta_0^{d_0-k}I_k^{(i)}\\
            &=-\frac{\bbE[Z]}{\delta_0(1-\delta_0)}+\frac{d_0(1-\delta_0) \bbE[Y]}{\delta_0(1-\delta_0)}\\
            &=\frac{-1}{\delta_0(1-\delta_0)}(\bbE[Z]-\bbE[X]\bbE[Y])\\
            &=\frac{-1}{\delta_0(1-\delta_0)}(\bbE[XY]-\bbE[X]\bbE[Y])\\
            &=\frac{-1}{\delta_0(1-\delta_0)}\bbE[(X-d_0(1-\delta_0))Y]\\
            &\leq \frac{-1}{\delta_0(1-\delta_0)}\bbE[(X-\lceil d_0(1-\delta_0)\rceil)Y]\\
            &=\frac{-1}{\delta_0(1-\delta_0)}\bbE[(X-\lceil d_0(1-\delta_0)\rceil)(Y-I_{\lceil d_0(1-\delta_0)\rceil}^{(i)})]\\
            &\overset{(a)}{\leq} 0
        \end{align}
        where (a) follows since $(k-\lceil d_0(1-\delta_0)\rceil)(I_k-I_{\lceil d_0(1-\delta_0)\rceil}^{(i)})\geq 0$. Thus, $\bbE[Y]$ decreases  $\delta_0$ increases.
Note that when $\delta_0$ increases to $1$, the upper bound \eqref{Eq: dropout MI ub} shrinks to $0$. 

Consider a special case when $I_k^{(i)}$ is proportional to $k$, i.e., $I_k^{(i)}=\alpha k$ for some constant $\alpha \in \bbR_{>0}$.  \eqref{Eq: dropout MI ub} can be rewritten as
\begin{align}
    &\sum_{e\in \{0,1\}^{d_0}} P_{E_t}(e) \sI(X_i \odot e,Y_i;\bW)\\
    &\leq \sum_{k=0}^{d_0} \binom{d_0}{k}\delta_0^{d_0-k}(1-\delta_0)^{k} \alpha k \\
    &= \alpha d_0(1-\delta_0)
\end{align}
which decays linearly in $\delta_0$.


\subsection{Proof of \cref{Lem: dropconnect SDPI coeff}}\label{App: pf of Lem: dropconnect SDPI coeff}
From \eqref{Eq: dropconnect layer}, for $i=1,\ldots,d_l$,  the $i\textsuperscript{th}$ element of $T_l$ is 
\begin{align}
    T_l(i)&=\activ_l\bigg(\sum_{j=1}^{d_{l-1}} \bW_l(i,j) \bE_l(i,j)T_{l-1}(j)\bigg)\\
    &=:\activ_l\bigg(\sum_{j=1}^{d_{l-1}} \bW_l(i,j)\widetilde{T}_{l-1}(i,j)\bigg),
\end{align}
where $\widetilde{T}_{l-1}$ is a $d_l \times d_{l-1}$ matrix.

Thus, given the network weights $\bW$, at any layer $l$, we have $d_l$ Markov chains: for $i=1,\ldots,d_l$, $T_{l-1}\to \widetilde{T}_{l-1}(i,:) \to T_l(i)$, where $\widetilde{T}_{l-1}(i,:)$ denotes the $i\textsuperscript{th}$ row of $\widetilde{T}_{l-1}$. We observe that for any $i\in [d_l]$, $P_{T_l(i)|\widetilde{T}_{l-1}(i,:),\bW}$ is a deterministic mapping, while $P_{\widetilde{T}_{l-1}(i,:)|T_{l-1}}$ is composed of $d_{l-1}$ parallel independet $\sZ$-channels, i.e., $P_{\widetilde{T}_{l-1}(i,:)|T_{l-1}}=\prod_{j=1}^{d_{l-1}}P_{\widetilde{T}|T}^{\delta_{l-1,i,j}}$, with $P_{\widetilde{T}|T}^{\delta_{l-1,i,j}}$ defined in \cref{App: pf of dropout}. Since each row of $\widetilde{T}_{l-1}$ are independent with each other given $T_{l-1}$, $P_{\widetilde{T}_{l-1}|T_{l-1}}=\prod_{i=1}^{d_l} P_{\widetilde{T}_{l-1}(i,:)|T_{l-1}}=\prod_{i=1}^{d_l}\prod_{j=1}^{d_{l-1}}P_{\widetilde{T}|T}^{\delta_{l-1,i,j}}$.

 The the following SDPI holds: for any training data sample index $i\in[n]$,
\begin{align}
    &\sD_\KL(P_{T_{l+1,i}|Y_i,\bW}\| P_{T_{l+1}|Y,\bW} | P_{Y_i,\bW})\\
    &\leq \bbE\big[\bbE[\eta_\KL(P_{T_{l+1}|\widetilde{T}_l,\bW}) \sD_\KL(P_{\widetilde{T}_{l,i}|Y_i,\bW}\| P_{\widetilde{T}_{l}|Y,\bW} ) | Y_i,\bW] \big]\\
    &\leq  \bbE\big[\bbE[\eta_\KL(P_{T_{l+1}|\widetilde{T}_l,\bW}) \eta_\KL(P_{\widetilde{T}_l|T_l}) \\
    &\qquad \times \sD_\KL(P_{T_{l,i}|Y_i,\bW}\| P_{T_{l}|Y,\bW} ) | Y_i,\bW] \big]\\
    &\overset{\text{(a)}}{=} \eta_\KL\bigg(\prod_{j=1}^{d_{l+1}}\prod_{k=1}^{d_{l}}P_{\widetilde{T}|T}^{\delta_{l,j,k}}\bigg) \sD_\KL(P_{T_{l,i}|Y_i,\bW}\| P_{T_{l}|Y,\bW} | P_{Y_i,\bW})\\
    &\overset{\text{(b)}}{\leq }  \bigg(1-\prod_{j=1}^{d_{l+1}}\bigg(1-\eta_\KL\bigg(\prod_{k=1}^{d_{l}}P_{\widetilde{T}|T}^{\delta_{l,j,k}} \bigg) \bigg) \bigg)\\
    &\qquad \times \sD_\KL(P_{T_{l,i}|Y_i,\bW}\| P_{T_{l}|Y,\bW} | P_{Y_i,\bW})
\end{align}
where (a) follows from $\eta_\KL(P_{T_{l+1}|\widetilde{T}_l,\bW})=1$ since $P_{T_{l+1}|\widetilde{T}_l,\bW}$ is a deterministic mapping, (b) follows from \cite[Theorem 5]{polyanskiy2017strong}. By applying \cite[Theorem 5]{polyanskiy2017strong} again, we have
\begin{align}
&\eta_\KL\bigg(\prod_{k=1}^{d_{l}}P_{\widetilde{T}|T}^{\delta_{l,j,k}} \bigg) \leq 1- \prod_{k=1}^{d_l} \delta_{l,j,k}, \quad  \text{and}\\
&1-\prod_{j=1}^{d_{l+1}}\bigg(1-\eta_\KL\bigg(\prod_{k=1}^{d_{l}}P_{\widetilde{T}|T}^{\delta_{l,j,k}} \bigg)  \leq 1-\prod_{j=1}^{d_{l+1}}\prod_{k=1}^{d_l} \delta_{l,j,k}.
\end{align}

\subsection{Proof of Lemma \ref{Lem: Y=g(x)+noise coeff}}\label{App: pf of Lem: Y=g(x)+noise coeff}

When $Y=g(X)+\epsilon N$, where $N$ is an independent noise and $\epsilon>0$ is a constant controlling the signal-to-noise ratio (SNR), then $P_{Y|X}=P_{g(X)+\epsilon N}$.  
The Dobrushin's coefficient is given by
\begin{align}
    \eta_{\TV}(P_{Y|X})=\sup_{x,x'\in\bbR^{d_x}}\|P_{g(x)+\epsilon N}-P_{g(x')+\epsilon N}\|_{\TV}. 
\end{align}
Let $N$ be a Gaussian noise generated from $\calN(\mathbf{0},\bI_{d_y})$. Denote the vector function $g(\cdot)$ by $g(\cdot)=(g_1(\cdot),\ldots,g_{d_y}(\cdot) )$.
Following the proof in \cite{polyanskiy2015dissipation} and the total variation distance between two Gaussians with the same covariance matrix in \cite[Theorem 1]{barsov1987estimates}, we have for any $x,x'\in\calX$,
\begin{align}
    &\sD_{\TV}(P_{g(x)+\epsilon N},P_{g(x')+\epsilon N})\\
    &=\| \calN(g(x),\epsilon^2 \bI_{d_y}) - \calN(g(x'),\epsilon^2 \bI_{d_y})\|_\TV\\
    &=1-2\qfunc\bigg(\frac{\|g(x)-g(x')\|}{2 \epsilon}\bigg)\\
    &= 1- 2\qfunc\bigg(\frac{\sqrt{\sum_{i=1}^{d_y}(g_i(x)-g_i(x'))^2}}{2 \epsilon}\bigg)\\
    &\leq 1- 2\qfunc\bigg(\frac{\sqrt{d_y(\|g(x)\|_\infty^2+\|g(x')\|_\infty^2)}}{2\epsilon}\bigg)\\
    &\leq 1-2\qfunc\bigg(\frac{\sqrt{2d_y}\|g\|_\infty}{2\epsilon}\bigg)
\end{align}
where $\qfunc(x)=\int_x^\infty \frac{1}{\sqrt{2\pi}}e^{-t^2/2} \rmd t$ is the Gaussian complimentary CDF. Finally, we have
\begin{align}
    \eta_f(P_{Y|X})\leq \eta_{\TV}(P_{Y|X}) \leq 1-2\qfunc\bigg(\frac{\sqrt{2d_y}\|g\|_\infty}{2\epsilon}\bigg).
\end{align}

\subsection{Proof of Theorem \ref{Thm: noisy DNN gen ub}}\label{App: pf of Thm: noisy DNN gen ub}

Among the commonly used activation functions, the following functions and their gradients are bounded: for any $u\in\bbR$,
\begin{itemize}[leftmargin=0.4cm]
    \item sigmoid function: $\mathrm{sigmoid}(u)=\frac{1}{1+e^{-u}}\in[0,1]$, $\mathrm{sigmoid}'(u)=\frac{e^{-u}}{(1+e^{-u})^2}\in[0,1]$. 
    \item softmax function: $\mathrm{softmax}(u)_i=\frac{e^{u_i}}{\sum_{j}e^{u_j}}\in[0,1]$, $\mathrm{softmax}'(u)_i=\mathrm{softmax}(u)_i(1-\mathrm{softmax}(u)_i)\in[0,1]$.
    \item $\tanh$ function: $\tanh{u}=\frac{e^u-e^{-u}}{e^u+e^{-u}}\in[-1,1]$, $\tanh'(u)=1-\tanh^2{u}\in[0,1]$.
\end{itemize}
As shown in \eqref{eq:noisy_DNN}, that is,  $\widetilde{T}_l=g_{\bW_l}(\widetilde{T}_{l-1})+\epsilon_l Z_l=\activ_l(\bW_l\widetilde{T}_{l-1})+\epsilon_l Z_l$ and from Lemma \ref{Lem: Y=g(x)+noise coeff}, the Dobrushin's coefficient at the $l\textsuperscript{th}$ layer is upper bounded by
\begin{align}
    \eta_{\TV}(P_{\widetilde{T}_l|\widetilde{T}_{l-1},\bW})&\leq  1-2\qfunc\bigg(\frac{\sqrt{2d_{l}}\|g_{\bW_l}\|_\infty}{2\epsilon_l}\bigg)\\
    &= 1-2\qfunc\bigg(\frac{\sqrt{2d_{l}}\|\activ_l\|_\infty}{2\epsilon_l}\bigg),
\end{align}
where $\|\activ_l\|_\infty=1$ for $\activ_l\in\{\mathrm{sigmoid}, \mathrm{softmax}, \tanh \}$.

In the following proof, at no risk of confusion, we let $T_l=\widetilde{T}_l$ and $T_{l,i}=\widetilde{T}_{l,i}$, for simplicity.
Conditioned on $\bW$, $P_{T_{l-1,i}|Y_i,\bW}\neq P_{T_{l-1}|Y,\bW}$ but $P_{T_{l,i}|T_{l-1,i},\bW}=P_{T_{l}|T_{l-1},W}$.  Thus, we have
\begin{small}
\begin{align}
    &\sD_\KL(P_{T_{l,i}|Y_i,\bW}\| P_{T_{l}|Y,\bW} | P_{Y_i,\bW})\\
    &\!\!\!=
    \sD_\KL(P_{T_{l,i}|T_{l-1,i},\bW} \!\circ\! P_{T_{l-1,i}|Y_i,\bW} \| P_{T_{l}|T_{l-1},W}\circ P_{T_{l-1}|Y,\bW} | P_{Y_i,\bW})\\
    &\leq \!\bigg(\! 1-2\qfunc\bigg(\frac{\sqrt{2d_{l}}\|\activ_l\|_\infty}{2\epsilon_l}\!\bigg)\!\!\bigg) \sD_\KL(P_{T_{l-1,i}|Y_i,\bW}\|P_{T_{l-1}|Y,\bW} | P_{Y_i,\bW}).
\end{align}
\end{small}
By induction, if we have $L$ layers
\begin{small}
\begin{align}
    &\sD_\KL(P_{T_{L,i}|Y_i,\bW}\| P_{T_{L}|Y,\bW} | P_{Y_i,\bW})\\
    &\leq \!\bigg(\! 1-2\qfunc\bigg(\frac{\sqrt{2d_{L}}\|\activ_L\|_\infty}{2\epsilon_L}\!\!\bigg)\!\!\bigg)  \sD_\KL(P_{T_{L-1,i}|Y_i,\bW}\| P_{T_{L-1}|Y,\bW} | P_{Y_i,\bW})\\
    &\vdots\\
    &\leq \prod_{l=1}^L\bigg(1-2\qfunc\bigg(\frac{\sqrt{2d_{l}}\|\activ_l\|_\infty}{2\epsilon_l} \bigg)\bigg)  \sD_\KL(P_{T_{0,i}|Y_i,\bW}\| P_{T_{0}|Y,\bW} | P_{Y_i,\bW})\\
    &=\prod_{l=1}^L\bigg(1-2\qfunc\bigg( \frac{\sqrt{2d_{l}}\|\activ_l\|_\infty}{2\epsilon_l} \bigg)\bigg) \sI(X_i;\bW|Y_i).
    \label{Eq: nosiy NN KL SDPI}
\end{align}
\end{small}

By combining \eqref{Eq: nosiy NN KL SDPI} with \eqref{Eq: noisy NN gen ub middle step}, the expected generalization error is upper bounded by
\begin{align}
    &| \gen(P_{\bW|D_n},P_{X,Y}) | \\
    &\!\!\! \leq\!\! \frac{\sqrt{2\sigma^2}}{n}\!\!\sum_{i=1}^n\!\sqrt{\prod_{l=1}^{L}\!\bigg(\! 1\!-\! 2\qfunc \!\bigg(\!\frac{\sqrt{2d_{l}}\|\activ_l\|_\infty}{2\epsilon_l}\!\!\bigg)\!\!\bigg) \sI(X_i;\bW|Y_i)\!+\!\sI(Y_i;\bW)}.
\end{align}
The proof is thus completed.

\section*{Acknowledgements} 	
The authors would like to sincerely thank the reviewers and the Associate Editor Prof.\ Varun Jog for many helpful comments to correct the subtle errors in the proofs and improve the presentation of the paper. 
Comments from Prof. Christina Lee Yu are also gratefully acknowledged.


\bibliographystyle{IEEEtran}
\bibliography{IEEEabrv,ref}

%

\begin{IEEEbiographynophoto}{Haiyun He (Member, IEEE)} was born in China in 1994.
 She received the B.E.\ degree from the Department of Electronics and Information Engineering at  Beihang University (BUAA) in 2016,  the M.Sc.\ degree in electrical engineering from the National University of Singapore (NUS) in 2017,  and the Ph.D.\ degree in electrical and computer engineering from the NUS in 2022.  
 She is currently a postdoctoral associate in the Center for Applied Mathematics at Cornell University.
 Her research lies at the intersection of information theory and machine learning (ML), focusing on developing fundamental theoretical analyses and effective practical solutions for ML challenges using information-theoretic tools. In 2022, she received the EECS Rising Star award from UT Austin.
\end{IEEEbiographynophoto}
\begin{IEEEbiographynophoto}{Ziv Goldfeld (Member, IEEE)}
received the B.Sc., M.Sc., and Ph.D. degrees
in electrical and computer engineering from Ben-Gurion University, Israel,
in 2012, 2014, and 2017, respectively. From 2017 to 2019, he was a
Post-Doctoral Fellow with the Laboratory for Information and Decision
Systems (LIDS), MIT. He is currently an Assistant Professor in the School of Electrical and
Computer Engineering at Cornell University.

He is the recipient of several awards, including the NSF CAREER Award, the IBM Academic Award, the Michael Tien '72 Excellence in  Teaching Award, and
the Rothschild Fellowship.
\end{IEEEbiographynophoto}

\end{document}